%% file: main.tex
\documentclass{egpubl}
\usepackage{eg2020}
\usepackage{csquotes}

\STAR                   %

\usepackage[T1]{fontenc}
\usepackage{dfadobe}  

\usepackage{subcaption}
\usepackage{cite}  %
\BibtexOrBiblatex
\electronicVersion
\PrintedOrElectronic
\ifpdf \usepackage[pdftex]{graphicx} \pdfcompresslevel=9
\else \usepackage[dvips]{graphicx} \fi

\usepackage{egweblnk} 

\usepackage{multirow}
\usepackage{booktabs} %
\usepackage{amsmath}
\usepackage{cleveref}
\usepackage{adjustbox}
\usepackage{array}
\usepackage[normalem]{ulem}
\usepackage{pifont}
\usepackage[dvipsnames]{xcolor}
\usepackage{amssymb}
\usepackage{amsfonts}
\usepackage{pdfcomment}
\usepackage{setspace}
\usepackage{eso-pic,graphicx}

\usepackage[absolute,overlay]{textpos}

\title[State of the Art on Neural Rendering]{State of the Art on Neural Rendering}

\author[A. Tewari \& O. Fried \& J. Thies et al.]
{
\parbox{\textwidth}
{\centering \vspace{-0.5cm}
A. Tewari$^{1\star}$~~O. Fried$^{2\star}$~~J. Thies$^{3\star}$~~V. Sitzmann$^{2\star}$~~S. Lombardi$^{4}$~~K. Sunkavalli$^{5}$~~R. Martin-Brualla$^{6}$~~T. Simon$^{4}$~~J. Saragih$^{4}$~~M. Nie{\ss}ner$^{3}$~~\\
R. Pandey$^{6}$~~S. Fanello$^{6}$~~G. Wetzstein$^{2}$~~J.-Y. Zhu$^{5}$~~C. Theobalt$^{1}$~~M. Agrawala$^{2}$~~E. Shechtman$^{5}$~~D. B Goldman$^{6}$~~M. Zollh{\"o}fer$^{4}$
}
\\
\parbox{\textwidth}
{\centering
$^{1}$MPI Informatics~~$^{2}$Stanford University~~$^{3}$Technical University of Munich~~$^{4}$Facebook Reality Labs~~$^{5}$Adobe Research~~$^{6}$Google Inc~~$^{\star}$Equal contribution.
}
}

\ifdefined\HIGHRES
    \graphicspath{{./applications/images/}}
\else
    \graphicspath{{./applications/images_lq/}}
\fi

\begin{document}

\begin{textblock*}{16cm}(2.4cm,0.4cm) 
   \noindent
   This is the accepted version of the following article: "State of the Art on Neural Rendering", which has been published in final form at \url{http://onlinelibrary.wiley.com}. This article may be used for non-commercial purposes in accordance with the Wiley Self-Archiving Policy [\url{http://olabout.wiley.com/WileyCDA/Section/id-820227.html}].
\end{textblock*}
\newcommand*{\ShowNotes}{}
\input{macros}

\teaser{
   \vspace{-0.75cm}
   \includegraphics[width=\linewidth]{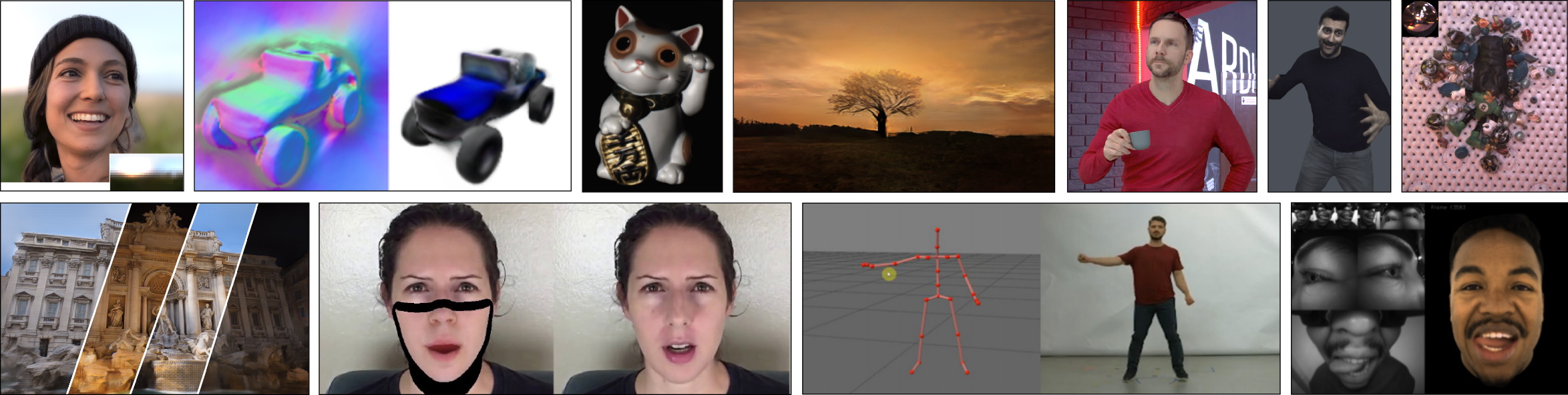}
   \centering
   \vspace{-0.5cm}
   \caption{
    Neural renderings of a large variety of scenes.
    See \Cref{sec:applications} for more details on the various methods.
    \newstuff{\footnotesize{Images from
    \cite{sun2019single, sitzmann2019srns, xu2019deepviewsynthesis, kar2017stereo, relightables, Martin-Brualla:2018:LEP:3272127.3275099, xu2018deeprelighting, Meshry_2019_CVPR, Fried2019, LiuBody2018, Wei:2019:VFA:3306346.3323030}}.
    }
   }
   \vspace{0.5cm}
   \label{fig:teaser}
}

\maketitle

\input{0_abstract}

\input{1_intro}

\input{2_related_surveys}

\input{3_scope}

\input{4_fundamentals}

\input{5_neural_rendering}

\input{6_applications}

\input{8_open_challenges}

\input{7_social_implications}

\input{9_conclusions}

\input{10_acks}

\bibliographystyle{eg-alpha-doi} 
\bibliography{main}

\newpage

\end{document}

%% file: macros.tex
\newcommand{\ignorethis}[1]{}
\newcommand{\redund}[1]{#1}

\newcommand{\etal       }     {{et~al.}}
\newcommand{\apriori    }     {\textit{a~priori}}
\newcommand{\aposteriori}     {\textit{a~posteriori}}
\newcommand{\perse      }     {\textit{per~se}}
\newcommand{\eg         }     {{e.g.~}}
\newcommand{\Eg         }     {{E.g.~}}
\newcommand{\ie         }     {{i.e.~}}
\newcommand{\naive      }     {{na\"{\i}ve}}

\newcommand{\Identity   }     {\mat{I}}
\newcommand{\Zero       }     {\mathbf{0}}
\newcommand{\Reals      }     {{\textrm{I\kern-0.18em R}}}
\newcommand{\isdefined  }     {\mbox{\hspace{0.5ex}:=\hspace{0.5ex}}}
\newcommand{\texthalf   }     {\ensuremath{\textstyle\frac{1}{2}}}
\newcommand{\half       }     {\ensuremath{\frac{1}{2}}}
\newcommand{\third      }     {\ensuremath{\frac{1}{3}}}
\newcommand{\fourth     }     {\ensuremath{\frac{1}{4}}}

\newcommand{\degree} {\ensuremath{^{\circ}}}

\newcommand{\mat        } [1] {{\text{\boldmath $\mathbit{#1}$}}}
\newcommand{\Approx     } [1] {\widetilde{#1}}
\newcommand{\change     } [1] {\mbox{{\footnotesize $\Delta$} \kern-3pt}#1}

\newcommand{\Order      } [1] {O(#1)}
\newcommand{\set        } [1] {{\lbrace #1 \rbrace}}
\newcommand{\floor      } [1] {{\lfloor #1 \rfloor}}
\newcommand{\ceil       } [1] {{\lceil  #1 \rceil }}
\newcommand{\inverse    } [1] {{#1}^{-1}}
\newcommand{\transpose  } [1] {{#1}^\mathrm{T}}
\newcommand{\invtransp  } [1] {{#1}^{-\mathrm{T}}}
\newcommand{\relu       } [1] {{\lbrack #1 \rbrack_+}}

\newcommand{\abs        } [1] {{| #1 |}}
\newcommand{\Abs        } [1] {{\left| #1 \right|}}
\newcommand{\norm       } [1] {{\| #1 \|}}
\newcommand{\Norm       } [1] {{\left\| #1 \right\|}}
\newcommand{\pnorm      } [2] {\norm{#1}_{#2}}
\newcommand{\Pnorm      } [2] {\Norm{#1}_{#2}}
\newcommand{\inner      } [2] {{\langle {#1} \, | \, {#2} \rangle}}
\newcommand{\Inner      } [2] {{\left\langle \begin{array}{@{}c|c@{}}
                               \displaystyle {#1} & \displaystyle {#2}
                               \end{array} \right\rangle}}

\newcommand{\twopartdef}[4]
{
  \left\{
  \begin{array}{ll}
    #1 & \mbox{if } #2 \\
    #3 & \mbox{if } #4
  \end{array}
  \right.
}

\newcommand{\fourpartdef}[8]
{
  \left\{
  \begin{array}{ll}
    #1 & \mbox{if } #2 \\
    #3 & \mbox{if } #4 \\
    #5 & \mbox{if } #6 \\
    #7 & \mbox{if } #8
  \end{array}
  \right.
}

\newcommand{\len}[1]{\text{len}(#1)}

\newlength{\w}
\newlength{\h}
\newlength{\x}

\definecolor{darkred}{rgb}{0.7,0.1,0.1}
\definecolor{darkgreen}{rgb}{0.1,0.5,0.1}
\definecolor{cyan}{rgb}{0.7,0.0,0.7}
\definecolor{otherblue}{rgb}{0.1,0.4,0.8}
\definecolor{maroon}{rgb}{0.76,.13,.28}
\definecolor{burntorange}{rgb}{0.81,.33,0}
\definecolor{othergreen}{rgb}{0.29,0.49,0.07}
\definecolor{orange}{rgb}{1.0,0.65,0.0}

\ifdefined\ShowNotes
  \newcommand{\colornote}[3]{{\color{#1}\textbf{#2 #3}}}
\else
  \newcommand{\colornote}[3]{}
\fi

\newcommand {\note}[1]{\colornote{maroon}{}{#1}}
\newcommand {\todo}[1]{\colornote{cyan}{TODO}{#1}}

\newcommand {\AT}[1]{\colornote{burntorange}{AT:}{#1}}
\newcommand {\ohad}[1]{\colornote{blue}{OF:}{#1}}
\newcommand {\justus}[1]{\colornote{otherblue}{JT:}{#1}}
\newcommand {\vincent}[1]{\colornote{darkgreen}{VS:}{#1}}
\newcommand {\stephen}[1]{\colornote{red}{SL:}{#1}}
\newcommand {\KS}[1]{\colornote{orange}{KS:}{#1}}
\newcommand {\ricardo}[1]{\colornote{brown}{RMB:}{#1}}
\newcommand {\tomas}[1]{\colornote{purple}{TS:}{#1}}
\newcommand {\jason}[1]{\colornote{gray}{JS:}{#1}}
\newcommand {\MATTHIAS}[1]{\colornote{red}{\textbf{MN:}{#1}}}
\newcommand {\rohit}[1]{\colornote{lime}{RP:}{#1}}
\newcommand {\SF}[1]{\colornote{magenta}{SF:}{#1}}
\newcommand {\junyanz}[1]{\colornote{orange}{JYZ:}{#1}}
\newcommand {\gordon}[1]{\colornote{pink}{GW:}{#1}}
\newcommand {\christian}[1]{\colornote{purple}{CT:}{#1}}
\newcommand {\eli}[1]{\colornote{teal}{ES:}{#1}}
\newcommand {\dan}[1]{\colornote{violet}{DG:}{#1}}
\newcommand {\maneesh}[1]{\colornote{yellow}{MA:}{#1}}
\newcommand {\MZ}[1]{\colornote{blue}{MZ:}{#1}}

\newcommand {\new}[1]{\colornote{red}{#1}}

\newcolumntype{R}[2]{%
    >{\adjustbox{angle=#1,lap=\width-(#2)}\bgroup}%
    l%
    <{\egroup}%
}
\newcommand*\rot{\multicolumn{1}{R{45}{1em}}}

\newcommand {\newstuff}[1]{#1}
\newcommand {\newnewstuff}[1]{#1}
\newcommand {\finalstuff}[1]{#1}

\newcommand\todosilent[1]{}

\newcommand{\woBGmask}{{w/o~bg~\&~mask}}
\newcommand{\woMask}{{w/o~mask}}

\definecolor{cmarkcolor}{rgb}{0.49,0.74,0.49}
\definecolor{xmarkcolor}{rgb}{0.86,0.34,0.34}
\newcommand{\cmark}{\textcolor{cmarkcolor}{\ding{51}}}
\newcommand{\xmark}{\textcolor{xmarkcolor}{\ding{55}}}

\definecolor{opA}{rgb}{0.88,0.55,0.11}
\definecolor{opB}{rgb}{0.70,0.43,0.65}
\definecolor{opC}{rgb}{0.34,0.42,0.72}
\definecolor{opD}{rgb}{0.49,0.74,0.49} %

\newcommand{\tableopt}[2]{{\color{#1}\textbf{#2}}}

\newcommand {\opA}[1]{\hyperlink{in_out_options}{\tableopt{opA}{#1}}}
\newcommand {\opB}[1]{\hyperlink{contents_options}{\tableopt{opB}{#1}}}
\newcommand {\opC}[1]{\hyperlink{control_options}{\tableopt{opC}{#1}}}
\newcommand {\opD}[1]{\hyperlink{cg_options}{\tableopt{opD}{#1}}}

\newcommand {\opE}[1]{\tableopt{burntorange}{#1}}
\newcommand {\opF}[1]{\tableopt{red}{#1}}
\newcommand {\opG}[1]{\tableopt{cyan}{#1}}
\newcommand {\opH}[1]{\tableopt{orange}{#1}}
\newcommand {\opI}[1]{\tableopt{magenta}{#1}}
\newcommand {\opJ}[1]{\tableopt{brown}{#1}}
\newcommand {\opK}[1]{\tableopt{gray}{#1}}
\newcommand {\opL}[1]{\tableopt{lightgray}{#1}}
\newcommand {\opM}[1]{\tableopt{lime}{#1}}
\newcommand {\opN}[1]{\tableopt{darkgray}{#1}}

\newcommand {\opO}[1]{\tableopt{olive}{#1}}
\newcommand {\opP}[1]{\tableopt{pink}{#1}}
\newcommand {\opQ}[1]{\tableopt{purple}{#1}}
\newcommand {\opR}[1]{\tableopt{teal}{#1}}
\newcommand {\opS}[1]{\tableopt{violet}{#1}}
\newcommand {\opT}[1]{\tableopt{violet}{#1}}

\newcommand {\rev}[1]{\tableopt{darkred}{#1}}

%% file: 0_abstract.tex
\begin{abstract}
Efficient rendering of photo-realistic virtual worlds is a long standing effort of computer graphics. Modern graphics techniques have succeeded in synthesizing photo-realistic images from hand-crafted scene representations.
However, the automatic generation of shape, materials, lighting, and other aspects of scenes  remains a challenging problem that, if solved, would make photo-realistic computer graphics more widely accessible.
Concurrently, progress in computer vision and machine learning have given rise to a new approach to image synthesis and editing, namely deep generative models.
Neural rendering is a new and rapidly emerging field that combines generative machine learning techniques with physical knowledge from computer graphics, e.g., by the integration of differentiable rendering into network training.
With a plethora of applications in computer graphics and vision, neural rendering is poised to become a new area in the graphics community, yet no survey of this emerging field exists.
This state-of-the-art report summarizes the recent trends and applications of neural rendering.
We focus on approaches that combine classic computer graphics techniques with deep generative models to obtain controllable and photo-realistic outputs.
Starting with an overview of the underlying computer graphics and machine learning concepts, we discuss critical aspects of neural rendering approaches.
Specifically, our emphasis is on the type of control, i.e., how the control is provided, which parts of the pipeline are learned, explicit vs.~implicit control, generalization, and stochastic vs.~deterministic synthesis.
The second half of this state-of-the-art report is focused on the many important use cases for the described algorithms such as novel view synthesis, semantic photo manipulation, facial and body reenactment, relighting, free-viewpoint video, and the creation of photo-realistic avatars for virtual and augmented reality telepresence.
Finally, we conclude with a discussion of the social implications of such technology and investigate open research problems.
\end{abstract}

%% file: 1_intro.tex
\section{Introduction}
\label{sec:intro}
\vspace{-0.1cm}
The creation of photo-realistic imagery of virtual worlds has been one of the primary driving forces for the development of sophisticated computer graphics techniques.
Computer graphics approaches span the range from real-time rendering, which enables the latest generation of computer games, to sophisticated global illumination simulation for the creation of photo-realistic digital humans in feature films.
In both cases, one of the main bottlenecks is content creation, i.e., that a vast amount of tedious and expensive manual work of skilled artists is required for the creation of the underlying scene representations in terms of surface geometry, appearance/material, light sources, and animations.
Concurrently, powerful generative models have emerged in the computer vision and machine learning communities.
The seminal work on \textit{Generative Adversarial Neural Networks} (GANs) by Goodfellow et al.~\cite{goodfellow2014} has evolved in recent years into deep generative models for the creation of high resolution imagery~\cite{radford2016unsupervised,Karras2017ProgressiveGO,brock2019large} and videos~\cite{vondrick2016generating,Clark2019}.
Here, control over the synthesized content can be achieved by conditioning \cite{pix2pix2016,zhu2017unpaired} the networks on control parameters or images from other domains.
Very recently, the two areas have come together and have been explored as ``neural rendering''.
One of the first publications that used the term neural rendering is \textit{Generative Query Network} (GQN) \cite{eslami2018neural}.
It enables machines to learn to perceive their surroundings based on a representation and generation network.
The authors argue that the network has an implicit notion of 3D due to the fact that it could take a varying number of images of the scene as input, and output arbitrary views with correct occlusion.
Instead of an implicit notion of 3D, a variety of other methods followed that include this notion of 3D more explicitly, exploiting components of the graphics pipeline.

While classical computer graphics starts from the perspective of physics, by modeling for example geometry, surface properties and cameras, machine learning comes from a statistical perspective, i.e., learning from real world examples to generate new images.
To this end, the quality of computer graphics generated imagery relies on the physical correctness of the employed models, while the quality of the machine learning approaches mostly relies on carefully-designed machine learning models and the quality of the used training data.
Explicit reconstruction of scene properties is hard and error prone and leads to artifacts in the  rendered content.
To this end, image-based rendering methods try to overcome these issues, by using simple heuristics to combine captured imagery.
But in complex scenery, these methods show artifacts like seams or ghosting.
Neural rendering brings the promise of addressing both \textit{reconstruction and rendering} by using deep networks to learn complex mappings from captured images to novel images.
Neural rendering combines physical knowledge, e.g., mathematical models of projection, with learned components to yield new and powerful algorithms for controllable image generation.
Neural rendering has not yet a clear definition in the literature.
Here, we define \textit{Neural Rendering} as: 
\begin{displayquote}
\textit{Deep image or video generation approaches that enable explicit or implicit control of scene properties such as illumination, camera parameters, pose, geometry, appearance, and semantic structure.}
\end{displayquote}
This state-of-the-art report defines and classifies the different types of neural rendering approaches.
Our discussion focuses on methods that combine computer graphics and learning-based primitives to yield new and powerful algorithms for \emph{controllable} image generation, since controllability in the image generation process is essential for many computer graphics applications.
One central scheme around which we structure this report is the kind of control afforded by each approach.
We start by discussing the fundamental concepts of computer graphics, vision, and machine learning that are prerequisites for neural rendering.
Afterwards, we discuss critical aspects of neural rendering approaches, such as: type of control, how the control is provided, which parts of the pipeline are learned, explicit vs.~implicit control, generalization, and stochastic vs.~deterministic synthesis. %
Following, we discuss the landscape of applications that is enabled by neural rendering.
The applications of neural rendering range from novel view synthesis, semantic photo manipulation, facial and body reenactment, relighting, free-viewpoint video, to the creation of photo-realistic avatars for virtual and augmented reality telepresence %
Since the creation and manipulation of images that are indistinguishable from real photos has many social implications, especially when humans are photographed, we also discuss these implications and the detectability of synthetic content.
As the field of neural rendering is still rapidly evolving, we conclude with current open research problems.

%% file: 2_related_surveys.tex
\vspace{-0.1cm}
\section{Related Surveys and Course Notes}

Deep Generative Models have been widely studied in the literature, with several surveys~\cite{salakhutdinov2015learning,Oussidi18,ou2018review} and course notes~\cite{openAIblog, StanfordCourse, IJCAICourse} describing them. 
Several reports focus on specific generative models, such as \textit{Generative Adversarial Networks} (GANs)~\cite{wang2019generative, creswell2018generative, goodfellow2016nips,CVPR2018tutorialGAN,Pan2018GANSurvey} and \textit{Variational Autoencoders} (VAEs)~\cite{doersch2016tutorial,kingma2019introduction}.
Controllable image synthesis using classic computer graphics and vision techniques have also been studied extensively.
Image-based rendering has been discussed in several survey reports~\cite{shum2000review,zhang2004survey}.
The book of Szeliski~\cite{szeliski2010computer} gives an excellent introduction to 3D reconstruction and image-based rendering techniques. 
Recent survey reports~\cite{egger20193d,Zollhoefer2018FaceSTAR} discuss approaches for 3D reconstruction and controllable rendering of faces for various applications.
Some aspects of neural rendering have been covered in tutorials and workshops of recent computer vision conferences.
These include approaches for free viewpoint rendering and relighting of full body performances~\cite{ECCV2018tutorialUltra,CVPR2018tutorialUltra, CVPR2019tutorialUltra}, tutorials on neural rendering for face synthesis~\cite{ECCV2018tutorialFace} and 3D scene generation using neural networks~\cite{CVPR2019tutorial3DSene}.
However, none of the above surveys and courses provide a structured and comprehensive look into neural rendering and all of its various applications.

%% file: 3_scope.tex
\vspace{-0.1cm}
\section{Scope of this STAR}

In this state-of-the-art report, we focus on novel approaches that combine classical computer graphics pipelines and learnable components.
Specifically, we are discussing where and how classical rendering pipelines can be improved by machine learning and which data is required for training.
To give a comprehensive overview, we also give a short introduction to the pertinent fundamentals of both fields, i.e., computer graphics and machine learning.
The benefits of the current hybrids are shown, as well as their limitations.
This report also discusses novel applications that are empowered by these techniques.
We focus on techniques with the primary goal of generating controllable photo-realistic imagery via machine learning.
We do not cover work on geometric and 3D deep learning \cite{mescheder2019occupancy,saito2019pifu,qi2016pointnet,choy20163d,park2019deepsdf}, which is more focused on 3D reconstruction and scene understanding. %
This branch of work is highly inspiring for many neural rendering approaches, especially ones that are based on 3D-structured scene representations, but goes beyond the scope of this survey.
We are also not focused on techniques that employ machine learning for denoising raytraced imagery \cite{Chaitanya:2017,LBF}.

%% file: 4_fundamentals.tex
\vspace{-0.1cm}
\section{Theoretical Fundamentals}
In the following, we discuss theoretical fundamentals of work in the neural rendering space.
First, we discuss image formation models in computer graphics, followed by classic image synthesis methods.
Next, we discuss approaches to generative models in deep learning.

\vspace{-0.1cm}
\subsection{Physical Image Formation}

Classical computer graphics methods approximate the physical process of image formation in the real world:
light sources emit photons that interact with the objects in the scene, as a function of their geometry and material properties, before being recorded by a camera. This process is known as light transport.
Camera optics acquire and focus incoming light from an aperture onto a sensor or film plane inside the camera body. The sensor or film records the amount of incident light on that plane, sometimes in a nonlinear fashion. 
All the components of image formation---light sources, material properties, and camera sensors---are wavelength-dependent. 
Real films and sensors often record only one to three different wavelength distributions, tuned to the sensitivity of the human visual system.
All the steps of this physical image formation are modelled in computer graphics:
light sources, scene geometry, material properties, light transport, optics, and sensor behavior.

\subsubsection{Scene Representations}
To model objects in a scene, many different representations for scene geometry have been proposed.
They can be classified into \textit{explicit} and \textit{implicit} representations.
Explicit methods describe scenes as a collection of geometric primitives, such as triangles, point-like primitives, or higher-order parametric surfaces.
Implicit representations include signed distance functions mapping from $\mathbb{R}^3 \rightarrow \mathbb{R}$, such that the surface is defined as the zero-crossing of the function (or any other level-set).
In practice, most hardware and software renderers are tuned to work best on triangle meshes, and will convert other representations into triangles for rendering.

The interactions of light with scene surfaces depend on the material properties of the surfaces.
Materials may be represented as bidirectional reflectance distribution functions (BRDFs) or bidirectional subsurface scattering reflectance distribution functions (BSSRDFs).
A BRDF is a 5-dimensional function that describes how much light of a given wavelength incident on a surface point from each incoming ray direction is reflected toward each exiting ray direction.
While a BRDF only models light interactions that happen at a single surface point, a BSSDRF models how light incident on one surface point is reflected at a different surface point, thus making it a 7-D function.
BRDFs can be represented using analytical models \cite{phong1975phong,cook1982reflectance,oren1995rough} or measured data \cite{matusik2003brdf}.
When a BRDF changes across a surface, it is referred to as a spatially-varying BRDF (svBRDF).
Spatially varying behavior across geometry may be represented by binding discrete materials to different geometric primitives, or via the use of texture mapping.
A texture map defines a set of continuous values of a material parameter, such as diffuse albedo, from a 2- or 3-dimensional domain onto a surface.
3-dimensional textures represent the value throughout a bounded region of space and can be applied to either explicit or implicit geometry.
2-dimensional textures map from a 2-dimensional domain onto a parametric surface; thus, they are typically applicable only to explicit geometry.

Sources of light in a scene can be represented using parametric models; these include point or directional lights, or area sources that are represented by surfaces in the scene that emit light. %
Some methods account for continuously varying emission over a surface, defined by a texture map or function.
Often environment maps are used to represent dense, distant scene lighting.
These environment maps can be stored as non-parametric textures on a sphere or cube, or can be approximated by coefficients of a spherical harmonic basis~\cite{Muller1966}.
Any of the parameters of a scene might be modeled as varying over time, allowing both animation across successive frames, and simulations of motion blur within a single frame.

\subsubsection{Camera Models}

The most common camera model in computer graphics is the pinhole camera model, in which rays of light pass through a pinhole and hit a film plane (image plane).
Such a camera can be parameterized by the pinhole's 3{D} location, the image plane, and a rectangular region in that plane representing the spatial extent of the sensor or film.
The operation of such a camera can be represented compactly using projective geometry, which converts 3D geometric representations using homogeneous coordinates into the two-dimensional domain of the image plane.
This is also known as a full perspective projection model.
Approximations of this model such as the weak perspective projection are often used in computer vision to reduce complexity because of the non-linearity of the full perspective projection.
More accurate projection models in computer graphics take into account the effects of non-ideal lenses, including distortion, aberration, vignetting, defocus blur, and even the inter-reflections between lens elements~\cite{Sturm:2011}.

\subsubsection{Classical Rendering}

The process of transforming a scene definition including cameras, lights, surface geometry and material into a simulated camera image is known as rendering.
The two most common approaches to rendering are rasterization and raytracing:
\textit{Rasterization} is a feed-forward process in which geometry is transformed into the image domain, sometimes in back-to-front order known as painter's algorithm.
\textit{Raytracing} is a process in which rays are cast backwards from the image pixels into a virtual scene, and reflections and refractions are simulated by recursively casting new rays from the intersections with the geometry~\cite{Whitted1980}. 

Hardware-accelerated rendering typically relies on rasterization, because it has good memory coherence.
However, many real-world image effects such as global illumination and other forms of complex light transport, depth of field, motion blur, etc. are more easily simulated using raytracing, and recent GPUs now feature acceleration structures to enable certain uses of raytracing in real-time graphics pipelines (e.g., NVIDIA RTX or DirectX Raytracing \cite{Haines2019}).
Although rasterization requires an explicit geometric representation, raytracing/raycasting can also be applied to implicit representations.
In practice, implicit representations can also be converted to explicit forms for rasterization using the marching cubes algorithm~\cite{Lorensen1987} and other similar methods.
Renderers can also use combinations of rasterization and raycasting to obtain high efficiency and physical realism at the same time (e.g., screen space ray-tracing~\cite{McGuire2014RayTrace}).
The quality of images produced by a given rendering pipeline depends heavily on the accuracy of the different models in the pipeline.
The components must account for the discrete nature of computer simulation, such as the gaps between pixel centers, using careful application of sampling and signal reconstruction theory.
The process of estimating the different model parameters (camera, geometry, material, light parameters) from real-world data, for the purpose of generating novel views, editing materials or illumination, or creating new animations is known as \textit{inverse rendering}.
\newstuff{Inverse rendering~\cite{marschner1998inverse,DADDB19,henzler2018escaping, deschaintre2018single,li2018materials   }}, which has been explored in the context of both computer vision and computer graphics, is closely related to neural rendering.
A drawback of inverse rendering is that the predefined physical \newstuff{model} or data structures used in classical rendering don't always accurately reproduce all the features of real-world physical processes, due to either mathematical complexity or computational expense.
In contrast, neural rendering introduces learned components into the rendering pipeline in place of such models.
Deep neural nets can statistically approximate such physical processes, resulting in outputs that more closely match the training data, reproducing some real-world effects more accurately than inverse rendering.

\newstuff{
Note that there are approaches at the intersection of inverse rendering and neural rendering.
E.g., Li et al.~\cite{li2018learning} uses a neural renderer that approximates global illumination effects to efficiently train an inverse rendering method that predicts depth, normal, albedo and roughness maps.
There are also approaches that use neural networks to enhance specific building blocks of the classical rendering pipeline, e.g., shaders.
Rainer et al.~\cite{Rainer19Neural} learn Bidirectional Texture Functions and Maximov et al.~\cite{Maximov_2019_ICCV} learn Appearance Maps.
}

\subsubsection{Light Transport}

Light transport considers all the possible paths of light from the emitting light sources, through a scene, and onto a camera. A well-known formulation of this problem is the classical rendering equation \cite{kajiya1986rendering}:

\begin{equation*}
L_{\text{o}}(\mathbf p,\, \omega_{\text{o}},\, \lambda,\, t) \,=\, L_{\text{e}}(\mathbf p,\, \omega_{\text{o}},\, \lambda,\, t) \ +\ L_{\text{r}}(\mathbf p,\, \omega_{\text{o}},\, \lambda,\, t)
\end{equation*}
where $L_{\text{o}}$ represents outgoing radiance from a surface as a function of location, ray direction, wavelength, and time.
The term $L_{\text{e}}$ represents direct surface emission, and the term $L_{\text{r}}$ represents the interaction of incident light with surface reflectance:
\begin{equation*}
    L_{\text{r}}(\mathbf p,\, \omega_{\text{o}},\, \lambda,\, t) \,=\, \int_\Omega f_{\text{r}}(\mathbf p,\, \omega_{\text{i}},\, \omega_{\text{o}},\, \lambda,\, t)\, L_{\text{i}}(\mathbf p,\, \omega_{\text{i}},\, \lambda,\, t)\, (\omega_{\text{i}}\,\cdot\,\mathbf n)\, \operatorname d \omega_{\text{i}}
\end{equation*}
Note that this formulation omits consideration of transparent objects and any effects of subsurface or volumetric scattering.
The rendering equation is an integral equation, and cannot be solved in closed form for nontrivial scenes, because the incident radiance $L_\text{i}$ appearing on the right hand side is the same as the outgoing radiance $L_\text{o}$ from another surface on the same ray. Therefore, a vast number of approximations have been developed. The most accurate approximations employ \textit{Monte Carlo} simulations\cite{veach1998thesis}, sampling ray paths through a scene. Faster approximations might expand the right hand side one or two times and then truncate the recurrence, thereby simulating only a few ``bounces'' of light. Computer graphics artists may also simulate additional bounces by adding non-physically based light sources to the scene.

\subsubsection{Image-based Rendering}
\label{fun:image_based_rendering}
In contrast to classical rendering, which projects 3D content to the 2D plane, image-based rendering techniques generate novel images by transforming an existing set of images, typically by warping and compositing them together.
Image-based rendering can handle animation, as shown by Thies et al.~\cite{thies2018headon}, but the most common use-case is novel view synthesis of static objects, in which image content from captured views are warped into a novel view based on a proxy geometry and estimated camera poses~\cite{Debevec1998,Gortler:1996:LUM:237170.237200,InsideOut2016}.
To generate a complete new image, multiple captured views have to be warped into the target view, requiring a blending stage.
The resulting image quality depends on the quality of the geometry, the number and arrangement of input views, and the material properties of the scene, since some materials change appearance dramatically across viewpoints.
Although heuristic methods for blending and the correction of view-dependent effects~\cite{InsideOut2016} show good results, recent research has substituted parts of these image-based rendering pipelines with learned components.
Deep neural networks have successfully been employed to reduce both blending artifacts~\cite{Hedman:2018:DBF:3272127.3275084} and artifacts that stem from view-dependent effects~\cite{thies2018ignor} (\Cref{app:neuralIBR}).

\vspace{-0.1cm}
\subsection{Deep Generative Models}
\label{fun:deep_generative_models}

While traditional computer graphics methods focus on physically modeling scenes and simulating light transport to generate images, machine learning can be employed to tackle this problem from a statistical standpoint, by learning the distribution of real world imagery. Compared to classical image-based rendering, which historically has used small sets of images (e.g., hundreds), deep generative models can learn image priors from large-scale image collections. 

Seminal work on deep generative models~\cite{ackley1985learning,hinton2006reducing,salakhutdinov2009deep} learned to generate random samples of simple digits and frontal faces. In these early results, both the quality and resolution was far from that achievable using physically-based rendering techniques. However, more recently, photo-realistic image synthesis has been demonstrated using \textit{Generative Adversarial Networks} (GANs)~\cite{goodfellow2014} and its extensions. Recent work can synthesize random high-resolution portraits that are often indistinguishable from real faces~\cite{karras2019style}. 

Deep generative models excel at generating \emph{random} realistic images with statistics resembling the training set. However, user control and interactivity play a key role in image synthesis and manipulation~\cite{barnes2009patchmatch}. %
For example, concept artists want to create particular scenes that reflect their design ideas rather than random scenes. Therefore, for computer graphics applications, generative models need to be extended to a conditional setting to gain explicit control of the image synthesis process. Early work trained feed-forward neural networks with a per-pixel $\ell_p$ distance to generate images given conditional inputs~\cite{dosovitskiy2015learning}. However, the generated results are often blurry as $\ell_p$ distance in pixel space considers each pixel independently and ignores the complexity of visual structure~\cite{pix2pix2016,blau2018perception}. Besides, it tends to average multiple possible outputs.
To address the above issue, recent work proposes perceptual similarity distances~\cite{gatys2016image,dosovitskiy2016generating,johnson2016perceptual} to measure the discrepancy between synthesized results and ground truth outputs in a high-level deep feature embedding space constructed by a pre-trained network. Applications include artistic stylization~\cite{gatys2016image,johnson2016perceptual}, image generation and synthesis~\cite{dosovitskiy2016generating,chen2017photographic}, and super-resolution~\cite{johnson2016perceptual,ledig2017photo}. %
Matching an output to its ground truth image does not guarantee that the output looks natural~\cite{blau2018perception}. Instead of minimizing the distance between outputs and targets, \textit{conditional GANs} (cGANs) aim to match the conditional distribution of outputs given inputs~\cite{MirzaO2014,pix2pix2016}. The results may not look the same as the ground truth images, but they look natural. Conditional GANs have been employed to bridge the gap between coarse computer graphics renderings and the corresponding real-world images~\cite{bi2019cg2real,zhu2017unpaired}, or to produce a realistic image given a user-specified semantic layout~\cite{pix2pix2016,park2019SPADE}. 
Below we provide more technical details for both network architectures and learning objectives. 

\vspace{-0.1cm}
\subsubsection{Learning a Generator}

We aim to learn a neural network $G$ that can map a conditional input $x \in \mathcal{X}$ to an output $y \in \mathcal{Y}$. Here $\mathcal{X}$ and $\mathcal{Y}$ denote the input and output domains. We call this neural network \emph{generator}. %
The conditional input $x$ can take on a variety of forms depending on the targeted application, such as a user-provided sketch image, camera parameters, lighting conditions, scene attributes, textual descriptions, among others. The output $y$ can also vary, from an image, a video, to 3D data such as voxels or meshes. See Table \ref{tbl:overview} for a complete list of possible network inputs and outputs for each application. 

Here we describe three commonly-used generator architectures. Readers are encouraged to check application-specific details in \Cref{sec:applications}. 
(1) \emph{Fully Convolutional Networks (FCNs)}~\cite{matan1992multi,long2015fully} are a family of models that can take an input image with arbitrary size and predict an output with the same size. Compared to popular image classification networks such as AlexNet~\cite{krizhevsky2012imagenet} and VGG~\cite{simonyan2015very} that map an image into a vector, FCNs use fractionally-strided convolutions to preserve the spatial image resolution~\cite{zeiler2010deconvolutional}. Although originally designed for recognition tasks such as semantic segmentation and object detection, FCNs have been widely used for many image synthesis tasks.
(2) \emph{U-Net}~\cite{10.1007/978-3-319-24574-4_28} is an FCN-based architecture with improved localization ability. The model adds so called ``skip connections'' from high-resolution feature maps at early layers to upsampled features in late-stage layers.  These skip connections help to produce more detailed outputs, since high-frequency information from the input can be passed directly to the output. (3) \emph{ResNet-based generators} use residual blocks ~\cite{he2016deep} to pass the high-frequency information from input to output, and have been used in style transfer~\cite{johnson2016perceptual} and image super-resolution~\cite{ledig2017photo}.%

\subsubsection{Learning using Perceptual Distance}

Once we collect many input-output pairs and choose a generator architecture, how can we learn a generator to produce a \emph{desired} output given an input? What would be an effective objective function for this learning problem? %
One straightforward way is to cast it as a regression problem, and to minimize the distance between the output $G(x)$ and its ground truth image $y$, as follows: 
\begin{equation}
\mathcal{L}_{\text{recon}}(G) = \mathbb{E}_{x, y} ||G(x) - y||_p,
\label{eqn:pixel_recon}
\end{equation}
where $\mathbb{E}$ denotes the expectation of the loss function over training pairs $(x, y)$, and $||\cdot||_p$ denotes the p-norm.  Common choices include $\ell_1$- or $\ell_2$-loss.
Unfortunately, the learned generator tends to synthesize blurry images or average results over multiple plausible outputs. For example, in image colorization, the learned generator sometimes produces desaturated results due to the averaging effect~\cite{zhang2016colorful}. In image super-resolution, the generator fails to synthesize structures and details as the p-norm looks at each pixel independently~\cite{johnson2016perceptual}.

To design a learning objective that better aligns with human's perception of image similarity, recent work~\cite{gatys2016image,johnson2016perceptual,dosovitskiy2016generating} proposes measuring the distance between deep feature representations extracted by a pre-trained image classifier $F$ (e.g., VGG network~\cite{simonyan2015very}). Such a loss is advantageous over the $\ell_p$-norm, as the deep representation summarizes an entire image holistically, while the $\ell_p$-norm evaluates the quality of each pixel independently. 
Mathematically, a generator is trained to minimize the following feature matching objective. 
\begin{equation}
\mathcal{L}_{\text{perc}}(G) = \mathbb{E}_{x, y} \sum_{t=1}^T \lambda_t \frac{1}{N_t}\big|\big|F^{(t)}(G(x))-F^{(t)}(y)\big|\big|_1,
\label{eqn:perceptual}
\end{equation}
where $F^{(t)}$ denotes the feature extractor in the $t$-th layer of the pre-trained network $F$ with $T$ layers in total and $N_t$ denoting the total number of features in layer $t$. The hyper-parameter $\lambda_t$ denotes the weight for each layer. 
Though the above distance is often coined ``perceptual distance'', it is intriguing why matching statistics in multi-level deep feature space can match human's perception and help synthesize higher-quality results, as the networks were originally trained for image classification tasks rather than image synthesis tasks.  A recent study~\cite{zhang2018unreasonable} suggests that rich features learned by strong classifiers also provide useful representations for human perceptual tasks, outperforming classic hand-crafted perceptual metrics~\cite{wang2004image,wang2003multiscale}.%

\subsubsection{Learning with Conditional GANs}
However, minimizing distances between output and ground truth does not guarantee realistic looking output,
according to the work of Blau and Michaeli~\cite{blau2018perception}. They also prove that the small distance and photorealism are at odds with each other. Therefore, instead of distance minimization, deep generative models focus on distribution matching, i.e., matching the distribution of generated results to the distribution of training data. Among many types of generative models, \textit{Generative Adversarial Networks} (GANs) have shown promising results for many computer graphics tasks. 
In the original work of Goodfellow et al.~\cite{goodfellow2014}, a GAN generator $G: z\rightarrow y$ learns a mapping from a low-dimensional random vector $z$ to an output image $y$. Typically, the input vector is sampled from a multivariate Gaussian or Uniform distribution. %
The generator $G$ is trained to produce outputs that cannot be distinguished from ``real'' images by an adversarially trained discriminator, $D$.
The discriminator is trained to detect synthetic images generated by the generator.
\newstuff{While GANs trained for object categories like faces or vehicles learn to synthesize high-quality instances of the object, usually the synthesized background is of a lower quality~\cite{karras2019style,Karras2017ProgressiveGO}.
Recent papers~\cite{Shaham_2019_ICCV,Ashual_2019_ICCV} try to alleviate this problem by learning generative models of a complete scene.}

To add conditional information as input, \textit{conditional GANs} (cGANs)~\cite{MirzaO2014,pix2pix2016} learn a mapping $G: \{x, z\} \rightarrow y$ from an observed input $x$ and a randomly sampled vector $z$ to an output image $y$. The observed input $x$ is also passed to the discriminator, which models whether image pairs $\{x, y\}$ are real or fake. As mentioned before, both input $x$ and output $y$ vary according to the targeted application. In class-conditional GANs~\cite{MirzaO2014},  the input $x$ is a categorical label that controls which object category a model should generate. %
In the case of image-conditional GANs such as \emph{pix2pix}~\cite{pix2pix2016}, the generator $G$ aims to translate an input image $x$, for example a semantic label map, to a realistic-looking output image, while the discriminator $D$ aims to distinguish real images from generated ones.
The model is trained with paired dataset $\{x_i,y_i\}_{i=1}^N$ that consists of pairs of corresponding input $x_i$ and output images $y_i$.
cGANs match the conditional distribution of the output given an input via the following minimax game:
\begin{equation}
\min_{G} \max_{D} \mathcal{L}_{\text{cGAN}}(G,D)\enspace{.}
\label{eqn::minimax}
\end{equation}
Here, the objective function $\mathcal{L}_{cGAN}(G,D)$
is normally defined as:
\begin{equation}
\mathcal{L}_{\text{cGAN}}(G, D) =
\mathbb{E}_{x, y} \big[ \!\log D(x, y) \big] +
\mathbb{E}_{x, z} \big[ \!\log\big(1 - D(x,G(x, z))\big) \big] \text{.}
\end{equation}
In early cGAN implementations~\cite{pix2pix2016,zhu2017unpaired}, no noise vector is injected, and the mapping is deterministic, as it tends to be ignored by the network during training. More recent work uses latent vectors $z$ to enable multimodal image synthesis~\cite{zhu2017toward,huang2018multimodal,almahairi2018augmented}. To stabilize training, cGANs-based methods~\cite{pix2pix2016,wang2018pix2pixHD} also adopt per-pixel $\ell_1$-loss $\mathcal{L}_{\text{recon}}(G)$ (\Cref{eqn:pixel_recon}) and perceptual distance loss $\mathcal{L}_{\text{perc}(G)}$ (\Cref{eqn:perceptual}).

During training, the discriminator $D$ tries to improve its ability to tell \emph{real} and \emph{synthetic} images apart, while the generator $G$, at the same time, tries to improve its capability of fooling the discriminator.
The pix2pix method adopts a U-Net~\cite{10.1007/978-3-319-24574-4_28} as the architecture of the generator and a patch-based fully convolutional network (FCN) ~\cite{long2015fully} as the discriminator.

Conceptually, perceptual distance and Conditional GANs are related, as both of them use an auxiliary network (either $F$ or $D$) to define an effective learning objective for learning a better generator $G$. In a high-level abstraction, an accurate computer vision model ($F$ or $D$) for assessing the quality of synthesized results $G(x)$ can significantly help tackle neural rendering problems. However, there are two significant differences. First, perceptual distance aims to measure the discrepancy between an output instance and its ground truth, while conditional GANs measure the closeness of the conditional distributions of real and fake images. Second, for perceptual distance, the feature extractor $F$ is pre-trained and fixed, while conditional GANs adapt its discriminator $D$ on the fly according to the generator. In practice, the two methods are complementary, and many neural rendering applications use both losses simultaneously~\cite{wang2018pix2pixHD,sungatullina2018image}.
Besides GANs, many promising research directions have recently emerged including Variational Autoencoders (VAEs)~\cite{jKingma2014}, auto-regressive networks (e.g., PixelCNN~\cite{Oord:2016}, PixelRNN~\cite{oord2016pixel,oord2016wavenet}), invertible density models~\cite{dinh2016density,kingma2018glow}, among others.
\newstuff{StarGAN \cite{STARGAN2018} enables training a single model for image-to-image translation based on multiple datasets with different domains.}
To keep the discussion concise, we focus on GANs here. We urge our readers to review tutorials~\cite{doersch2016tutorial,kingma2019introduction} and course notes~\cite{openAIblog, StanfordCourse, IJCAICourse} for a complete picture of deep generative models.

\subsubsection{Learning without Paired Data}
Learning a generator with the above objectives requires hundreds to millions of paired training data. In many real-world applications, paired training data are difficult and expensive to collect. Different from labeling images for classification tasks, annotators have to label every single pixel for image synthesis tasks.  For example, only a couple of small datasets exist for tasks like semantic segmentation. Obtaining input-output pairs for graphics tasks such as artistic stylization can be even more
challenging since the desired output often requires artistic authoring and is sometimes not even well-defined. 
In this setting, the model is given a source set $\{x_i\}_{i=1}^N$ ($x_i \in \mathcal{X}$) and a target set $\{y_j\}_{j=1}$ ($y_j \in \mathcal{Y}$).  All we know is which target \emph{domain} the output $G(x)$ should come from: i.e., like an image from domain $Y$. But given a particular input, we do not know which target \emph{image} the output should be.  There could be infinitely many mappings to project an image from $\mathcal{X}$ to $\mathcal{Y}$. Thus, we need additional constraints. Several constraints have been proposed including cycle-consistency loss for enforcing a bijective mapping~\cite{zhu2017unpaired,yi2017dualgan,kim2017learning}, the distance preserving loss for encouraging that the output is close to the input image either in pixel space~\cite{shrivastava2017learning} or in feature embedding space~\cite{bousmalis2017unsupervised,taigman2017unsupervised}, the weight sharing strategy for learning shared representation across domains~\cite{liu2016coupled,liu2017unsupervised,huang2018multimodal}, etc. The above methods broaden the application scope of conditional GANs and enable many graphics applications such as object transfiguration, domain transfer, and CG2real.%

%% file: 5_neural_rendering.tex
\section{Neural Rendering}

Given high-quality scene specifications, classic rendering methods can render photorealistic images for a variety of complex real-world phenomena.
Moreover, rendering gives us explicit editing control over all the elements of the scene---camera viewpoint, lighting, geometry and materials. 
However, building high-quality scene models, especially directly from images, requires significant manual effort, and automated scene modeling from images is an open research problem.
On the other hand, deep generative networks are now starting to produce visually compelling images and videos either from random noise, or conditioned on certain user specifications like scene segmentation and layout.
However, they do not yet allow for fine-grained control over scene appearance and cannot always handle the complex, non-local, 3D interactions between scene properties.
In contrast, neural rendering methods hold the promise of combining these approaches to enable controllable, high-quality synthesis of novel images from input images/videos. 
Neural rendering techniques are diverse, differing in the control they provide over scene appearance, the inputs they require, the outputs they produce, and the network structures they utilize. 
A typical neural rendering approach takes as input images corresponding to certain scene conditions (for example, viewpoint, lighting, layout, etc.), builds a ``neural'' scene representation from them, and ``renders'' this representation under novel scene properties to synthesize novel images. 
The learned scene representation is not restricted by simple scene modeling approximations and can be optimized for high quality novel images.
At the same time, neural rendering approaches incorporate ideas from classical graphics---in the form of input features, scene representations, and network architectures---to make the learning task easier, and the output more controllable.

We propose a taxonomy of neural rendering approaches along the axes that we consider the most important:
\begin{itemize}
    \item \textit{Control:} What do we want to control and how do we condition the rendering on the control signal?
    \item \textit{CG Modules:} Which computer graphics modules are used and how are they integrated into a neural rendering pipeline?
    \item \textit{Explicit or Implicit Control:} Does the method give explicit control over the parameters or is it done implicitly by showing an example of what we expect to get as output?
    \item \textit{Multi-modal Synthesis:} Is the method trained to output multiple optional outputs, given a specific input?
    \item \textit{Generality:} Is the rendering approach generalized over multiple scenes/objects?
\end{itemize}
In the following, we discuss these axes that we use to classify current state-of-the-art methods (see also \Cref{tbl:overview}).

\vspace{-0.1cm}
\subsection{Control}

Neural rendering aims to render high-quality images under user-specified scene conditions.
In the general case, this is an open research problem.
Instead, current methods tackle specific sub-problems like novel view synthesis~\cite{Hedman:2018:DBF:3272127.3275084,thies2018ignor,sitzmann2019deepvoxels,sitzmann2019srns}, relighting under novel lighting~\cite{xu2018deeprelighting,relightables}, and animating faces~\cite{Kim:2018:DVP:3197517.3201283,Thies:2019:DNR:3306346.3323035,Fried2019} and bodies~\cite{AbermanSLLCC19,Shysheya2019TexturedNA,Martin-Brualla:2018:LEP:3272127.3275099} under novel expressions and poses.
A main axis in which these approaches differ is in how the control signal is provided to the network.
One strategy is to directly pass the scene parameters as input to the first or an intermediate network layer~\cite{eslami2018neural}.
Related strategies are to tile the scene parameters across all pixels of an input image, or concatenate them to the activations of an inner network layer \cite{meka2019deep,sun2019single}.
Another approach is to rely on the spatial structure of images and employ an image-to-image translation network to map from a ``guide image'' or ``conditioning image'' to the output.
For example, such approaches might learn to map from a semantic mask to the output image~\cite{karacan2016learning,park2019SPADE,wang2018pix2pixHD,zhu2016generative,Bau:Ganpaint:2019,brock2017neural,chen2017photographic,pix2pix2016}.
Another option, which we describe in the following, is to use the control parameters as input to a graphics layer.

\vspace{-0.1cm}
\subsection{Computer Graphics Modules}
One emerging trend in neural rendering is the integration of computer graphics knowledge into the network design.
Therefore, approaches might differ in the level of ``classical'' graphics  knowledge that is embedded in the system.
For example, directly mapping from the scene parameters to the output image does not make use of any graphics knowledge.
One simple way to integrate graphics knowledge is a non-differentiable computer graphics module.
Such a module can for example be used to render an image of the scene and pass it as dense conditioning input to the network \cite{Kim:2018:DVP:3197517.3201283,LiuBody2018,Fried2019,Martin-Brualla:2018:LEP:3272127.3275099}.
Many different channels could be provided as network inputs, such as a depth map, normal map, camera/world space position maps, albedo map, a diffuse rendering of the scene, and many more.
This transforms the problem to an image-to-image translation task, which is a well researched setting, that can for example be tackled by a deep conditional generative model with skip connections.
A deeper integration of graphics knowledge into the network is possible based on differentiable graphics modules.
Such a differentiable module can for example implement a complete computer graphics renderer \cite{Lombardi:2019:NVL:3306346.3323020,sitzmann2019srns}, a 3D rotation \cite{sitzmann2019deepvoxels,nguyen2018rendernet,hologan}, or an illumination model \cite{shu2017neural}.
Such components add a physically inspired inductive bias to the network, while still allowing for end-to-end training via backpropagation.
This can be used to analytically enforce a truth about the world in the network structure, frees up network capacity, and leads to better generalization, especially if only limited training data is available.

\vspace{-0.1cm}
\subsection{Explicit vs. Implicit Control}
Another way to categorize neural rendering approaches is by the type of  control. 
Some approaches allow for explicit control, i.e., a user can edit the scene parameters manually in a semantically meaningful manner.
For example, current neural rendering approaches allow for explicit control over camera viewpoint~\cite{xu2019deepviewsynthesis,thies2018ignor,hologan,eslami2018neural,Hedman:2018:DBF:3272127.3275084,Aliev2019,Meshry_2019_CVPR,nguyen2018rendernet,sitzmann2019srns,sitzmann2019deepvoxels}, scene illumination~\cite{zhou2019portraitrelighting,xu2018deeprelighting,philip2019multiviewrelighting,meka2019deep,sun2019single}, facial pose and expression~\cite{Lombardi:2018:DAM:3197517.3201401,Thies:2019:DNR:3306346.3323035,Wei:2019:VFA:3306346.3323030,Kim:2018:DVP:3197517.3201283,Geng:2018:WGS:3272127.3275043}.
Other approaches only allow for implicit control by way of a representative sample.
While they can copy the scene parameters from a reference image/video, one cannot manipulate these parameters explicitly. 
This includes methods that transfer human head motion from a reference video to a target person~\cite{zakharov2019few}, or methods which retarget full-body motion~\cite{AbermanSLLCC19,Chan2018}
Methods which allow for explicit control require training datasets with images/videos and their corresponding scene parameters.%
On the other hand, implicit control usually requires less supervision. 
These methods can be trained without explicit 3D scene parameters, only with weaker annotations. 
For example, while dense facial performance capture is required to train networks with explicit control for facial reenactment~\cite{Kim:2018:DVP:3197517.3201283,Thies:2019:DNR:3306346.3323035}, implicit control can be achieved by training just on videos with corresponding sparse 2D keypoints~\cite{zakharov2019few}.  

\vspace{-0.1cm}
\subsection {Multi-modal Synthesis}

Often times it is beneficial to have several different output options to choose from.
For example, when only a subset of scene parameters are controlled, there potentially exists a large multi-modal output space with respect to the other scene parameters.
Instead of being presented with one single output, the user can be presented with a gallery of several choices, which are visibly different from each other. Such a gallery helps the user better understand the output landscape and pick a result to their liking. %
To achieve various outputs which are significantly different from each other the network or control signals must have some stochasticity or structured variance. For example, variational auto-encoders~\cite{jKingma2014,larsen2016autoencoding} model processes with built-in variability, and can be used to achieve multi-modal synthesis~\cite{walker2016uncertain,xue2016visual,zhu2017toward}. The latest example is Park~et~al.~\cite{park2019SPADE}, which demonstrates one way to incorporate variability and surfaces it via a user interface: given the same semantic map, strikingly different images are generated with the push of a button.

\vspace{-0.1cm}
\subsection {Generality}

Neural rendering methods differ in their object specificity. Some methods aim to train a general purpose model once, and apply it to all instances of the task at hand~\cite{xu2019deepviewsynthesis,sitzmann2019srns,nguyen2018rendernet,hologan,Hedman:2018:DBF:3272127.3275084,eslami2018neural,Bau:Ganpaint:2019,park2019SPADE,zhu2016generative,brock2017neural,zakharov2019few,pix2pix2016,karacan2016learning,chen2017photographic,wang2018pix2pixHD}. For example, if the method operates on human heads, it will aim to be applicable to all humans. Conversely, other methods are instance-specific~\cite{Chan2018,LiuBody2018,Lombardi:2018:DAM:3197517.3201401,Wei:2019:VFA:3306346.3323030,AbermanSLLCC19,sitzmann2019deepvoxels,Lombardi:2019:NVL:3306346.3323020,Kim:2018:DVP:3197517.3201283,Fried2019,thies2018ignor,Aliev2019,Meshry_2019_CVPR,sitzmann2019srns}. Continuing our human head example, these networks will operate on a single person (with a specific set of clothes, in a specific location) and a new network will have to be retrained for each new subject. For many tasks, object specific approaches are currently producing higher quality results, at the cost of lengthy training times for each object instance. For real-world applications such training times are prohibitive---improving general models is an open problem and an exciting research direction. %

%% file: 6_applications.tex
\vspace{-0.1cm}
\section{Applications of Neural Rendering}
\label{sec:applications}

\input{applications/tables/overview_table.tex}
Neural rendering has many important use cases such as semantic photo manipulation, novel view synthesis, relighting,  free-viewpoint video, as well as facial and body reenactment.
\Cref{tbl:overview} provides an overview of various applications discussed in this survey.
For each, we report the following attributes:
\begin{itemize}
\hypertarget{in_out_options}{
  \item \textit{\textbf{Required Data.}} All the data that is required for the system. This does not include derived data, e.g., automatically computed facial landmarks, but instead can be thought of as the minimal amount of data a person would have to acquire in order to be able to reproduce the system. 
}
  \item \textit{\textbf{Network Inputs.}} The data that is directly fed into the learned part of the system, i.e., the part of the system through which the gradients flow during backpropagation.
  \item \textit{\textbf{Network Outputs.}} Everything produced by the learned parts of the system. This is the last part of the pipeline in which supervision is provided.

    \vspace{0.1cm}
    {\small
    Possible values for \emph{Required Data}, \emph{Network Inputs} and \emph{Network Outputs}: 
    \opA{I}mages, \opA{V}ideos, \opA{M}eshes, \opA{N}oise, \opA{T}ext, \opA{C}amera, \opA{L}ighting, 2D \opA{J}oint positions, \opA{R}enders, \opA{S}emantic labels, 2D \opA{K}eypoints, volum\opA{E}, te\opA{X}tures, \opA{D}epth (for images or video).
    }
    \vspace{0.2cm}
    
\hypertarget{contents_options}{
  \item \textit{\textbf{Contents.}} The types of objects and environments that the system is designed to handle as input and output. 
  {\small
  Possible values: \opB{H}ead, \opB{P}erson, \opB{R}oom, outdoor \opB{E}nvironment, \opB{S}ingle object (of any category).
  }
}
\hypertarget{control_options}{  
  \item \textit{\textbf{Controllable Parameters.}} The parameters of the scene that can be modified. 
  {\small 
  Possible values: \opC{C}amera, \opC{P}ose, \opC{L}ighting, colo\opC{R}, \opC{T}exture, \opC{S}emantics, \opC{E}xpression, speec\opC{H}.
  }
}
  \item \textit{\textbf{Explicit control.}} Refers to systems in which the user is given interpretable parameters that, when changed, influence the generated output in a predictable way. 
  {\small
  Possible values: \xmark{} uninterpretable or uncontrollable, \cmark{} interpretable controllable parameters. 
  }
\hypertarget{cg_options}{   
  \item \textit{\textbf{CG module.}} The level of ``classical'' graphics knowledge embedded in the system. 
  {\small
  Possible values: \xmark{} no CG module, \opD{N}on-differentiable CG module, \opD{D}ifferentiable CG module.
  }
}
  \item \textit{\textbf{Generality.}} General systems are trained once and can be applied to multiple different input instances. \Eg a system that synthesizes humans, but has to be retrained for each new person, does not have this property.
  {\small
  Possible values: \xmark{} instance specific, \cmark{} general.
  }
  \item \textit{\textbf{Multi-modal synthesis.}} Systems that, as presented, allow on-demand generation of multiple outputs which are significantly different from each other, based on the same input.
  {\small
  Possible values: \xmark{} single output, \cmark{} on-demand multiple outputs.
  }
  \item \textit{\textbf{Temporal coherence.}} Specifies whether temporal coherence is explicitly enforced during training of the approach.
  {\small 
  Possible values: \xmark{} not enforced, \cmark{} enforced (e.g. in loss function).
  }
\end{itemize}
The following is a detailed discussion of various neural rendering applications.

\subsection{Semantic Photo Synthesis and Manipulation}
\label{app:semantic}

\input{applications/semantic_synthesis.tex}

\input{applications/semantic_editing.tex}

\input{applications/novel_view_synthesis.tex}

\input{applications/free_viewpoint_videos.tex}

\input{applications/relighting.tex}

\input{applications/facial_reenactment.tex}

\input{applications/neural-bodies.tex}

%% file: applications/tables/overview_table.tex
\begin{table*}
    \centering
    \small
    \begin{tabular}{llllllllllll@{}}
        \toprule
        Method &
        \rot{Required Data} &             
        \rot{Network Inputs} &            
        \rot{Network Outputs} &           
        \rot{Contents} &                  
        \rot{Controllable Parameters} &   
        \rot{Explicit Control} &         %
        \rot{CG Module} &                 %
        \rot{Generality} &               %
        \rot{Multi-modal Synthesis} &    %
        \rot{Temporal Coherence} \\        %

        \midrule
        Bau~\etal~\cite{Bau:Ganpaint:2019}
        & \opA{I}\opA{S}   &   \opA{I}\opA{S}  & \opA{I}      
        &  \opB{R}\opB{E} & \opC{S}    & \xmark{} & \xmark{} & \cmark{} & \xmark{} & \xmark{}         
        & \multirow[t]{9}{3cm}{Semantic Photo Synthesis (\Cref{app:semantic})} \\
        
        \cmidrule(lr){1-11}
        Brock~\etal~\cite{brock2017neural}
        & \opA{I}   &  \opA{N} & \opA{I} 
        & \opB{S} & \opC{R}      & \cmark{} & \xmark{} & \cmark{} & \xmark{} & \xmark{}      
        & \\
        
        \cmidrule(lr){1-11}
        Chen and Koltun~\cite{chen2017photographic}
        & \opA{I}\opA{S} &  \opA{S}  & \opA{I}          
        &  \opB{R}\opB{E} & \opC{S}       & \xmark{} & \xmark{} & \cmark{} & \cmark{} &  \xmark{}     
        & \\
        
        \cmidrule(lr){1-11} 
        Isola~\etal~\cite{pix2pix2016}
        & \opA{I}\opA{S}            &   \opA{S}  & \opA{I}          
        & \opB{E}\opB{S} & \opC{S}       & \xmark{} & \xmark{} & \cmark{} & \xmark{} & \xmark{}    
        & \\
        
        \cmidrule(lr){1-11}
         Karacan~\etal~\cite{karacan2016learning}
        & \opA{I}\opA{S}            &   \opA{S}  & \opA{I}          
        & \opB{E} & \opC{S}      & \xmark{} & \xmark{} & \cmark{} & \cmark{} & \xmark{}
        & \\

        \cmidrule(lr){1-11}
        Park~\etal~\cite{park2019SPADE}
        & \opA{I}\opA{S} &  \opA{S}  & \opA{I}          
        & \opB{R}\opB{E} & \opC{S}       
        & \xmark{} & \xmark{} & \cmark{} & \cmark{} & \xmark{} & \\
        
        \cmidrule(lr){1-11}
        Wang~\etal~\cite{wang2018pix2pixHD}
        & \opA{I}\opA{S} &  \opA{S}  & \opA{I}          
        & \opB{R}\opB{E}\opB{S} & \opC{S}       & \xmark{} & \xmark{} & \cmark{} & \cmark{} 
        & \xmark{} & \\
        
        \cmidrule(lr){1-11}
        Zhu~\etal~\cite{zhu2016generative}
        & \opA{I}   &   \opA{N} & \opA{I} 
        & \opB{E}\opB{S} & \opC{R}\opC{T}       & \cmark{} & \xmark{} & \cmark{} & \cmark{} & \xmark{}     
        & \\

        \midrule Aliev~\etal~\cite{Aliev2019}
        & \opA{I}\opA{D} & \opA{R} & \opA{I} & \opB{R}\opB{S} & \opC{C} & \cmark{} & \opD{N} & \xmark{} & \xmark{} & \xmark{} & 
        \multirow[t]{9}{3cm}{Novel View Synthesis (\Cref{subsec:novel_view})} \\    
        \cmidrule(lr){1-11} Eslami~\etal~\cite{eslami2018neural}
        & \opA{I}\opA{C}        &  \opA{I}\opA{C}          & \opA{I}               & \opB{R}\opB{S}               & \opC{C}        & \cmark{} & \xmark{} & \cmark{} & \xmark{} & \xmark{} & \\

        \cmidrule(lr){1-11} Hedman~\etal~\cite{Hedman:2018:DBF:3272127.3275084}
        & \opA{V} & \opA{I} & \opA{I} & \opB{R}\opB{E}\opB{S} & \opC{C} & \cmark{} & \opD{N} & \cmark{} & \xmark{} & \xmark{} & \\

        \cmidrule(lr){1-11} Meshry~\etal~\cite{Meshry_2019_CVPR} &
        \opA{I} & \opA{I}\opA{L} & \opA{I} & \opB{R}\opB{E} & \opC{C}\opC{L} & \cmark{} & \opD{N} & \xmark{} & \c mark{} & \xmark{} & \\
        
        \cmidrule(lr){1-11} Nguyen-Phuoc~\etal~\cite{nguyen2018rendernet}
        & \opA{I}\opA{C}\opA{L} & \opA{E} & \opA{I} & \opB{S} & \opC{C}\opC{L} & \cmark{} & \opD{N} & \cmark{} & \xmark{} & \xmark{} & \\

        \cmidrule(lr){1-11} Nguyen-Phuoc~\etal~\cite{hologan}
        & \opA{I} & \opA{N}\opA{C} & \opA{I} & \opB{S} & \opC{C} & \cmark{} & \xmark{} & \cmark{} & \cmark{} & \xmark{} & \\

        \cmidrule(lr){1-11} Sitzmann~\etal~\cite{sitzmann2019deepvoxels}
        & \opA{V}        & \opA{I}\opA{C}          & \opA{I}               & \opB{S}               & \opC{C}        & \cmark{} & \opD{D}  & \xmark{} & \xmark{} &  \xmark{}   & \\
        
        \cmidrule(lr){1-11} Sitzmann~\etal~\cite{sitzmann2019srns}
        & \opA{I}\opA{C}        & \opA{I}\opA{C}          & \opA{I}               & \opB{S}               & \opC{C}        & \cmark{} & \opD{D}  & \cmark{} & \xmark{} &  \xmark{}  & \\
        
        \cmidrule(lr){1-11} Thies~\etal~\cite{thies2018ignor}
        & \opA{V} & \opA{I}\opA{R}\opA{C} & \opA{I} & \opB{S} & \opC{C} & \cmark{} & \opD{N} & \xmark{} & \xmark{} & \xmark{} & \\
        
        \cmidrule(lr){1-11} Xu~\etal~\cite{xu2019deepviewsynthesis}
        & \opA{I}\opA{C}        & \opA{I}\opA{C} & \opA{I}   & \opB{S}               & \opC{C}        & \cmark{} & \xmark{} & \cmark{} & \xmark{} &  \xmark{}  & \\

        \midrule Lombardi~\etal~\cite{Lombardi:2019:NVL:3306346.3323020}
        & \opA{V}\opA{C}        & \opA{I}\opA{C}          & \opA{I}               & \opB{H}\opB{P}\opB{S} & \opC{C} & \cmark{} & \opD{D}  & \xmark{} & \xmark{} & \xmark{} & \multirow[t]{4}{3cm}{Free Viewpoint Video (\Cref{subsec:free_viewpoint})} \\
        
        \cmidrule(lr){1-11} Martin-Brualla~\etal~\cite{Martin-Brualla:2018:LEP:3272127.3275099}
        & \opA{V}\opA{D}\opA{C} & \opA{R}                 & \opA{V}               & \opB{P}        & \opC{C}        & \cmark{} & \opD{N} & \cmark{} & \xmark{} &        \cmark{}  & \\
        
        \cmidrule(lr){1-11} Pandey~\etal~\cite{Pandey_2019_CVPR} 
        & \opA{V}\opA{D}\opA{I}               & \opA{I}\opA{D}\opA{C}        & \opA{I}               & \opB{P}               & \opC{C} & \cmark{} & \xmark{} & \cmark{} & \xmark{} & \xmark{} & \\
        
        \cmidrule(lr){1-11} Shysheya~\etal~\cite{Shysheya2019TexturedNA}
        & \opA{V}        & \opA{R}             & \opA{I}               & \opB{P}               & \opC{C}\opC{P} & \cmark{} & \xmark{} & \cmark{} & \xmark{} & \xmark{} & \\

        \midrule Meka~\etal~\cite{meka2019deep} 
        & \opA{I}\opA{L}        & \opA{I}\opA{L}        & \opA{I}               & \opB{H}            & \opC{L}        & \cmark{} & \xmark{} & \cmark{} & \xmark{} & \xmark{} &
        \multirow[t]{4}{3cm}{Relighting \\ (\Cref{subsec:relighting})} \\

        \cmidrule(lr){1-11} Philip~\etal~\cite{philip2019multiviewrelighting}
        & \opA{I}        & \opA{I}\opA{L}  & \opA{I} & \opB{E}               & \opC{L}        & \cmark{} & \opD{N}  & \cmark{} & \xmark{} & \xmark{} & \\
        
        \cmidrule(lr){1-11} Sun~\etal~\cite{sun2019single}
        & \opA{I}\opA{L}        & \opA{I}\opA{L}        & \opA{I}\opA{L}        & \opB{H}               & \opC{L}        & \cmark{} & \xmark{} & \cmark{} & \xmark{} & \xmark{} & \\
        
        \cmidrule(lr){1-11} Xu~\etal~\cite{xu2018deeprelighting}
        & \opA{I}\opA{L}        & \opA{I}\opA{L} & \opA{I}               & \opB{S}              & \opC{L}        & \cmark{} & \xmark{} & \cmark{} & \xmark{} & \xmark{} & \\

        \cmidrule(lr){1-11} Zhou~\etal~\cite{zhou2019portraitrelighting}
        & \opA{I}\opA{L}        & \opA{I}\opA{L}        & \opA{I}\opA{L}        & \opB{H}               & \opC{L}        & \cmark{} & \xmark{} & \cmark{} & \xmark{} &  \xmark{} & \\

        \midrule Fried~\etal~\cite{Fried2019}
        & \opA{V}\opA{T}        & \opA{V}\opA{R}          & \opA{V}               & \opB{H}   & \opC{H}        & \cmark{} & \opD{N}  & \xmark{} & \xmark{} & \cmark{} & 
        \multirow[t]{6}{3cm}{Facial Reenactment (\Cref{subsec:face_body})} \\
        
        \cmidrule(lr){1-11} Kim~\etal~\cite{Kim:2018:DVP:3197517.3201283}
        & \opA{V}               & \opA{R}          & \opA{V}               & \opB{H}        & \opC{P}\opC{E} & \cmark{} & \opD{N}  & \xmark{} & \xmark{} &  \cmark{}        & \\
        
        \cmidrule(lr){1-11} Lombardi~\etal~\cite{Lombardi:2018:DAM:3197517.3201401} 
        & \opA{V}\opA{C} & \opA{I}\opA{M}\opA{C} & \opA{M}\opA{X}        & \opB{H}               & \opC{C}\opC{P} & \cmark{} & \opD{N} & \xmark{} & \xmark{} & \xmark{} & \\
        
        \cmidrule(lr){1-11} Thies~\etal~\cite{Thies:2019:DNR:3306346.3323035}
        &  \opA{V} & \opA{I}\opA{R}\opA{C} & \opA{I} & \opB{H}\opB{S} &  \opC{C}\opC{E} & \cmark{} & \opD{D} & \xmark{} & \xmark{} & \xmark{} & \\
        
        \cmidrule(lr){1-11} Wei~\etal~\cite{Wei:2019:VFA:3306346.3323030}
        & \opA{V}\opA{C} & \opA{I}                 & \opA{M}\opA{X} & \opB{H}               & \opC{C}\opC{P} & \cmark{} & \opD{D}  & \xmark{} & \xmark{} &    \xmark{}      & \\

        \cmidrule(lr){1-11} Zakharov~\etal~\cite{zakharov2019few}
        & \opA{I} & \opA{I}\opA{K} & \opA{I} & \opB{H} & \opC{P}\opC{E} & \xmark{} & \xmark{} & \cmark{} & \xmark{} & \xmark{} & \\

        \midrule Aberman~\etal~\cite{AbermanSLLCC19}
        & \opA{V}               & \opA{J}          & \opA{V}               & \opB{P}        & \opC{P}        & \xmark{} & \xmark{} & \xmark{} & \xmark{} & \cmark{} &
        \multirow[t]{3}{3cm}{Body Reenactment (\Cref{subsec:face_body})} \\
        
        \cmidrule(lr){1-11} Chan~\etal~\cite{Chan2018}
        & \opA{V}               & \opA{J}     & \opA{V}               & \opB{P}        & \opC{P}        & \xmark{} & \xmark{} & \xmark{} & \xmark{} & \cmark{} & \\
        
        \cmidrule(lr){1-11} Liu~\etal~\cite{LiuBody2018}
        & \opA{V}\opA{M}        & \opA{R}          & \opA{V}               & \opB{P}        & \opC{P}        & \cmark{} & \opD{N}  & \xmark{} & \xmark{} & \cmark{} & \\

        \midrule
        \noalign{\vskip 0.01cm}  
        \cmidrule(lr){2-5} \cmidrule(lr){6-7} \cmidrule(lr){8-11}
        & \multicolumn{4}{c}{Inputs and Outputs} & \multicolumn{2}{c}{Control} & \multicolumn{4}{c}{Misc} & \\
        \noalign{\vskip 0.01cm}
        \bottomrule
    \end{tabular}
    
    \caption{Selected methods presented in this survey. See \Cref{sec:applications} for explanation of attributes in the table and their possible values.
    }
    \label{tbl:overview}
\end{table*}

%% file: applications/semantic_synthesis.tex
\emph{Semantic photo synthesis and manipulation} enable interactive image editing tools for controlling and modifying the appearance of a photograph in a semantically meaningful way. The seminal work \emph{Image Analogies}~\cite{hertzmann2001image} creates new texture given a semantic layout and a reference image, using patch-based texture synthesis~\cite{efros1999texture,efros2001image}. Such single-image patch-based methods~\cite{hertzmann2001image,wexler07spacetime,simakov08bidir,barnes2009patchmatch} enable image reshuffling, retargeting, and inpainting, but they cannot allow high-level operations such as adding a new object or synthesizing an image from scratch. Data-driven graphics systems create new imagery by compositing multiple image regions~\cite{perez2003poisson} from images retrieved from a large-scale photo collection~\cite{lalonde2007photo,chen2009sketch2photo,johnson2006semantic,hays2007scene,Eitz:2009:PhotoSketch}. These methods allow the user to specify a desired scene layout using inputs such as a sketch~\cite{chen2009sketch2photo} or a semantic label map~\cite{johnson2006semantic}. The latest development is \emph{OpenShapes}~\cite{bansal2019shapes}, which composes regions by matching scene
context, shapes, and parts.  While achieving appealing results, these systems are often slow as they search in a large image database. In addition, undesired artifacts can be sometimes spotted due to visual inconsistency between different images. 
 \input{applications/figures/semantic_synthesis.tex}
 
\subsubsection{Semantic Photo Synthesis}
\label{app:semantic-synthesis}

In contrast to previous non-parametric approaches, recent work has trained fully convolutional networks~\cite{long2015fully} with a conditional GANs objective~\cite{MirzaO2014,pix2pix2016} to directly map a user-specified semantic layout to a photo-realistic image~\cite{pix2pix2016,karacan2016learning,liu2017unsupervised,zhu2017unpaired,yi2017dualgan,huang2018multimodal,wang2018pix2pixHD}. Other types of user inputs such as color, sketch, and texture have also been supported~\cite{sangkloy2017scribbler,iizuka2016let,zhang2016colorful,xian2018texturegan}. 
Among these, \emph{pix2pix}~\cite{pix2pix2016} and the method of Karacan \etal~\shortcite{karacan2016learning} present the first learning-based methods for semantic image synthesis including generating street view and natural scene images. To increase the image resolution, \emph{Cascaded refinement networks}~\cite{chen2017photographic} learn a coarse-to-fine generator, trained with a perceptual loss~\cite{gatys2016image}. The results are high-resolution, but lack high frequency texture and details. %
To synthesize richer details, \emph{pix2pixHD} ~\cite{wang2018pix2pixHD} proposes a conditional GAN that can generate $2048 \times 1024$ results with realistic texture. The key extensions compared to \emph{pix2pix}\cite{pix2pix2016} include a coarse-to-fine generator similar to CRN~\cite{chen2017photographic},  multi-scale discriminators that capture local image statistics at different scales, and a multi-scale discriminator-based feature matching objective, that resembles perceptual distance~\cite{gatys2016image}, but uses an adaptive discriminator to extract task-specific features instead. Notably, the multi-scale pipeline, a decades-old scheme in vision and graphics~\cite{burt1983laplacian,brown2003recognising}, is still highly effective for deep image synthesis. Both \emph{pix2pixHD}
and \emph{BicycleGAN}~\cite{zhu2017toward} can synthesize multiple possible outputs given the same user input, allowing a user to choose different styles. Subsequent systems~\cite{wang2018video,bansal2018recycle,bashkirova2018unsupervised} extend to the video domain, allowing a user to control the semantics of a video. Semi-parametric systems~\cite{qi2018semi,bansal2019shapes} combine classic data-driven image compositing~\cite{lalonde2007photo} and feed-forward networks~\cite{long2015fully}. %

Most recently, \emph{GauGAN}~\cite{park2019SPADE,park2019GauGAN} uses a SPatially-Adaptive (DE)normalization layer (SPADE) to better preserve semantic information in the generator. 
While previous conditional models~\cite{pix2pix2016,wang2018pix2pixHD} process a semantic layout through multiple normalization layers (e.g., InstanceNorm~\cite{ulyanov2016instance}), the channel-wise normalization layers tend to ``wash away'' semantic information, especially for uniform and flat input layout regions. Instead, the GauGAN generator takes a random latent vector as an image style code, and employs multiple ResNet blocks with spatially-adaptive normalization layers (SPADE), %
to produce the final output. As shown in \Cref{fig:semantic_synthesis}, this design not only produces visually appealing results, but also enables better user control over style and semantics. The adaptive normalization layers have also been found to be effective for stylization~\cite{huang2017arbitrary} and super-resolution~\cite{wang2018recovering}.

%% file: applications/figures/semantic_synthesis.tex
\begin{figure}[t]
	\centering
  \includegraphics[width=0.9\linewidth]{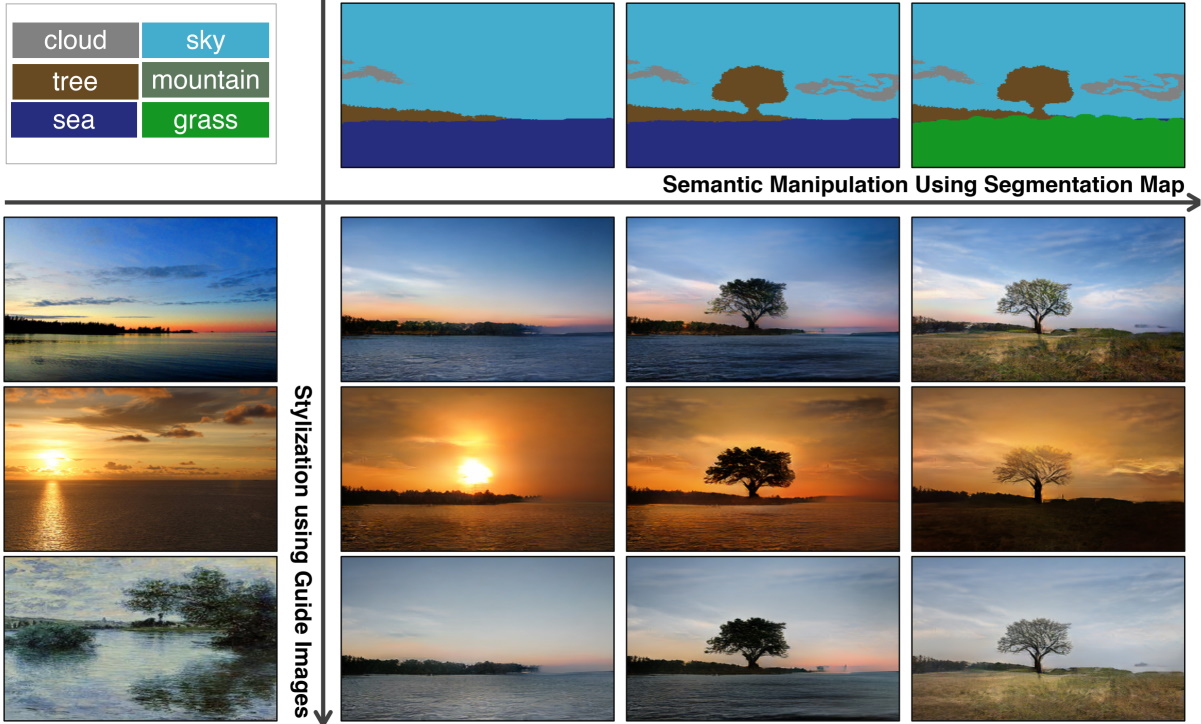}
\caption{\emph{GauGAN}~\cite{park2019SPADE,park2019GauGAN} enables image synthesis with both semantic and style control. Please see the SIGGRAPH 2019 Real-Time \href{https://youtu.be/Gz9weuemhDA?t=2949}{Live} for more details. 
\newstuff{\footnotesize{Images taken from Park et al.~\cite{park2019SPADE}}}.}
  \label{fig:semantic_synthesis}
\vspace{-0.6cm}
\end{figure}

%% file: applications/semantic_editing.tex
\input{applications/figures/semantic_editing.tex}
\subsubsection{Semantic Image Manipulation}
\label{app:semantic-editing}
The above image synthesis systems excel at creating new visual content, given user controls as inputs. However, \emph{semantic image manipulation} of a user provided image with deep generative models remains challenging for two reasons. First, editing an input image requires accurately reconstructing it with the generator, which is a difficult task even with recent GANs.
Second, once the controls are applied, the newly synthesized content might not be compatible with the input photo.
To address these issues, \emph{iGAN}~\cite{zhu2016generative} proposes using an unconditional GAN as a natural image prior for image editing tasks. The method first optimizes a low-dimensional latent vector such that the GAN can faithfully reproduce an input photo. The reconstruction method combines quasi-Newton optimization with encoder-based initialization. The system then modifies the appearance of the generated image using color, sketch, and warping tools. To render the result, they transfer the edits from the generated image to the original photo using guided image filtering~\cite{he2012guided}. 
Subsequent work on \emph{Neural Photo Editing}~\cite{brock2017neural} uses a VAE-GAN~\cite{larsen2016autoencoding} to encode an image into a latent vector and generates an output by blending the modified content and the original pixels. The system allows semantic editing of faces, such as adding a beard. Several works~\cite{perarnau2016invertible,yao20183d,hong2018learning} train an encoder together with the generator. They deploy a second encoder to predict additional image attributes (e.g., semantics, 3D information, facial attributes) and allow a user to modify these attributes. This idea of using GANs as a deep image prior was later used in image inpainting~\cite{yeh2017semantic} and deblurring~\cite{asim2018blind}.
The above systems work well on a low-resolution image with a single object or of a certain class and often require post-processing (e.g., filtering and blending) as the direct GANs' results are not realistic enough. To overcome these challenges, \emph{GANPaint}~\cite{Bau:Ganpaint:2019} adapts a pre-trained GAN model to a particular image. The learned image-specific GAN combines the prior learned from the entire image collection and image statistics of that particular image. Similar to prior work~\cite{zhu2016generative,brock2017neural}, the method first projects an input image into a latent vector. The reconstruction from the vector is close to the input, but many visual details are missing. 
The method then slightly changes the network's internal parameters to reconstruct more precisely the input image. During test time, GANPaint modifies intermediate representations of GANs~\cite{bau2019gandissect} according to user inputs. 
Instead of training a randomly initialized CNN on a single image as done in \emph{Deep Image Prior}~\cite{ulyanov2018deep}, \emph{GANPaint} leverages the prior learned from a pre-trained generative model and fine-tunes it for each input image.  
As shown in Figure~\ref{fig:semantic_editing}, this enables addition and removal of certain objects  in a realistic manner.
\newstuff{Learning distribution priors via pre-training, followed by fine-tuning on limited data, 
is useful for many One-shot and Few-shot synthesis scenarios \cite{10.5555/3326943.3327138,liu2019few}.}

\subsubsection{Improving the Realism of Synthetic Renderings}
\label{app:cg2real}

The methods discussed above use deep generative models to either synthesize images from user-specified semantic layouts, or modify a given input image in a semantically meaningful manner. As noted before, rendering methods in computer graphics have been developed for the exact same goal---generating photorealistic images from scene specifications. However, the visual quality of computer rendered images depends on the fidelity of the scene modeling; using low-quality scene models and/or rendering methods results in images that look obviously synthetic.
Johnson et al. \cite{johnson2011cg2real} addressed this issue by improving the realism of synthetic renderings using content from similar, real photographs retrieved from a large-scale photo collection. However, this approach is restricted by the size of the database and the simplicity of the matching metric.
Bi et al.~\cite{bi2019cg2real} propose using deep generative models to accomplish this task. They train a conditional generative model to translate a low-quality rendered image (along with auxiliary information like scene normals and diffuse albedo) to a high-quality photorealistic image. They propose performing this translation on an albedo-shading decomposition (instead of image pixels) to ensure that textures are preserved.
Shrivastava et al.~\cite{shrivastava2017learning} learn to improve the realism of renderings of the human eyes based on unlabeled real images and Mueller et al.~\cite{GANeratedHands_CVPR2018} employs a similar approach for human hands. Hoffman et al.~\cite{hoffman2018cycada} extends CycleGAN~\cite{zhu2017unpaired} with feature matching to improve the realism of street view rendering for domain adaptation. 
Along similar lines, Nalbach et al. \cite{Nalbach2017b} propose using deep convolutional networks to convert shading buffers such as per-pixel positions, normals, and material parameters into complex shading effects like ambient occlusion, global illumination, and depth-of-field, thus significantly speeding up the rendering process. The idea of using coarse renderings in conjunction with deep generative models to generate high-quality images has also been used in approaches for facial editing \cite{Kim:2018:DVP:3197517.3201283,Fried2019}.

%% file: applications/figures/semantic_editing.tex
\begin{figure}[t]
\centering
\includegraphics[width=\linewidth]{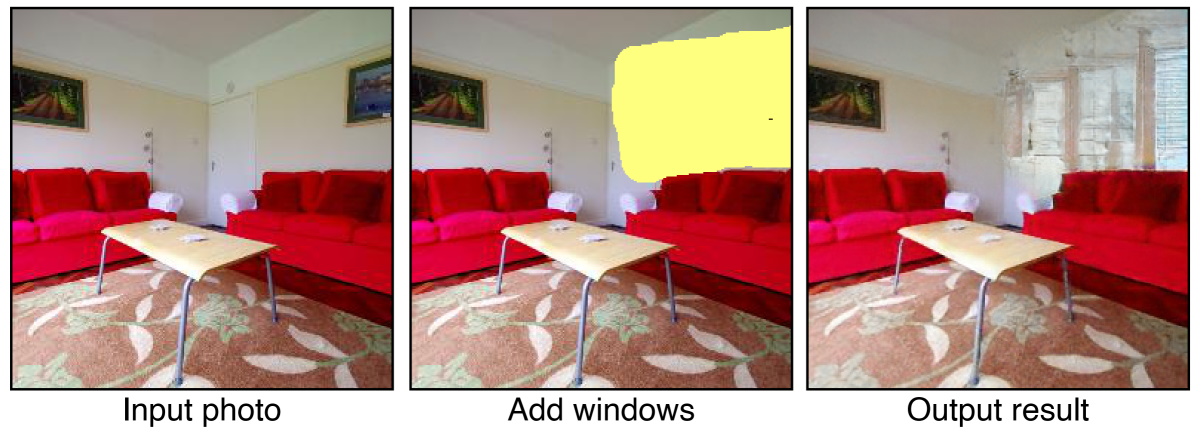}
\caption{\emph{GANPaint}\cite{Bau:Ganpaint:2019} enables a few high-level image editing operations. A user can add, remove, or alter an object in an image with simple brush tools.  A deep generative model will then satisfy user's constraint while preserving natural image statistics. 
\newstuff{\footnotesize{Images taken from Bau et al.~\cite{Bau:Ganpaint:2019}}}.}
\label{fig:semantic_editing}
\vspace{-0.35cm}
\end{figure}

%% file: applications/novel_view_synthesis.tex
\vspace{-0.1cm}
\subsection{Novel View Synthesis for Objects and Scenes}
\label{subsec:novel_view}

\emph{Novel view synthesis} is the problem of generating novel camera perspectives of a scene given a fixed set of images of the same scene. 
Novel view synthesis methods thus deal with image and video synthesis conditioned on camera pose.
Key challenges underlying novel view synthesis are inferring the scene's 3D structure given sparse observations, as well as inpainting of occluded and unseen parts of the scene.
In classical computer vision, image-based rendering (IBR) methods~\cite{Debevec1998,Chaurasia:2013:DSL:2487228.2487238} typically rely on optimization-based multi-view stereo methods to reconstruct scene geometry and warp observations into the coordinate frame of the novel view.
However, if only few observations are available, the scene contains view-dependent effects, or a large part of the novel perspective is not covered by the observations, IBR may fail, leading to results with ghosting-like artifacts and holes.
Neural rendering approaches have been proposed to generate higher quality results. 
In Neural Image-based Rendering~\cite{Hedman:2018:DBF:3272127.3275084,Meshry_2019_CVPR}, previously hand-crafted parts of the IBR pipeline are replaced or augmented by learning-based methods.
Other approaches reconstruct a learned representation of the scene from the observations, learning it end-to-end with a differentiable renderer. This enables learning of priors on geometry, appearance and other scene properties in a learned feature space.
Such neural scene representation-based approaches range from prescribing little structure on the representation and the renderer~\cite{eslami2018neural}, to proposing 3D-structured representations such as voxel grids of features~\cite{sitzmann2019srns, hologan}, to explicit 3D disentanglement of voxels and texture~\cite{zhu2018visual}, point clouds~\cite{Meshry_2019_CVPR}, multi-plane images~\cite{xu2019deepviewsynthesis,Flynn19DeepView}, or implicit functions~\cite{saito2019pifu,sitzmann2019srns} which equip the network with inductive biases on image formation and geometry.
Neural rendering approaches have made significant progress in previously open challenges such as the generation of view-dependent effects~\cite{thies2018ignor, xu2019deepviewsynthesis} or learning priors on shape and appearance from extremely sparse observations~\cite{sitzmann2019srns, xu2019deepviewsynthesis}.
While neural rendering shows better results compared to classical approaches, it still has limitations.
I.e., they are restricted to a specific use-case and are limited by the training data.
Especially, view-dependent effects such as reflections are still challenging.

\input{applications/neural-IBR.tex}

\input{applications/neural_rerendering_in_the_wild.tex}
\input{applications/mpi_based.tex}

\input{applications/neural_scene_reps_and_rendering.tex}
\input{applications/voxel_based.tex}
\input{applications/implicit_based.tex}

%% file: applications/neural-IBR.tex
\subsubsection{Neural Image-based Rendering}
\label{app:neuralIBR}

Neural Image-based Rendering (N-IBR) is a hybrid between classical image-based rendering and deep neural networks that replaces hand-crafted heuristics with learned components.
A classical IBR method uses a set of captured images and a proxy geometry to create new images, e.g., from a different viewpoint.
The proxy geometry is used to reproject image content from the captured images to the new target image domain.
In the target image domain, the projections from the source images are blended to composite the final image.
This simplified process gives accurate results only for diffuse objects with precise geometry reconstructed with a sufficient number of captured views.
However, artifacts such as ghosting, blur, holes, or seams can arise due to view-dependent effects, imperfect proxy geometry or too few source images.
To address these issues, N-IBR methods replace the heuristics often found in classical IBR methods with learned blending functions or corrections that take into account view-dependent effects.
\textit{DeepBlending}~\cite{Hedman:2018:DBF:3272127.3275084} proposes a generalized network to predict blending weights of the projected source images for compositing in the target image space.
They show impressive results on indoor scenes with fewer blending artifacts than classical IBR methods.
In \textit{Image-guided Neural Object Rendering}~\cite{thies2018ignor}, a scene specific network is trained to predict view-dependent effects with a network called \textit{EffectsNet}.
It is used to remove specular highlights from the source images to produce diffuse-only images, which can be projected into a target view without copying the source views' highlights.
This EffectsNet is trained in a Siamese fashion on two different views at the same time, enforcing a multi-view photo consistency loss.
In the target view, new view-dependent effects are reapplied and the images are blended using a U-Net-like architecture.
As a result, this method demonstrates novel view point synthesis on objects and small scenes including view-dependent effects.

%% file: applications/neural_rerendering_in_the_wild.tex
\input{applications/figures/neural-rerendering-in-the-wild.tex}

\subsubsection{Neural Rerendering}
\label{app:neural-rerendering-in-the-wild}

Neural Rerendering combines classical 3D representation and renderer with deep neural networks that rerender the classical render into a more complete and realistic views. In contrast to Neural Image-based Rendering (N-IBR), neural rerendering does not use input views at runtime, and instead relies on the deep neural network to recover the missing details. \textit{Neural Rerendering in the Wild}~\cite{Meshry_2019_CVPR} uses neural rerendering to synthesize realistic views of tourist landmarks under various lighting conditions, see \Cref{fig:neural-wild}. The authors cast the problem as a multi-modal image synthesis problem that takes as input a rendered deep buffer, containing depth and color channels, together with an appearance code, and outputs realistic views of the scene. The system reconstructs a dense colored point cloud from internet photos using Structure-from-Motion and Multi-View Stereo, and for each input photo, renders the recovered point cloud into the estimated camera. Using pairs of real photos and corresponding rendered deep buffers, a multi-modal image synthesis pipeline learns an implicit model of appearance, that represents time of day, weather conditions and other properties not present in the 3D model. To prevent the model from synthesizing transient objects, like pedestrians or cars, the authors propose conditioning the rerenderer with a semantic labeling of the expected image. At inference time, this semantic labeling can be constructed to omit any such transient objects. Pittaluga et al.~\cite{pittaluga2019revealing} use a neural rerendering technique to invert Structure-from-Motion reconstructions and highlight the privacy risks of Structure-from-Motion 3D reconstructions, that typically contain color and SIFT features. The authors show how sparse point clouds can be inverted and generate realistic novel views from them. In order to handle very sparse inputs, they propose a visibility network that classifies points as visible or not and is trained with ground-truth correspondences from 3D reconstructions.

%% file: applications/figures/neural-rerendering-in-the-wild.tex
\begin{figure}[t]
    \centering
    \captionsetup[subfigure]{aboveskip=1pt,belowskip=1pt}
    \begin{subfigure}[b]{0.49\linewidth}
        \includegraphics[width=\linewidth]{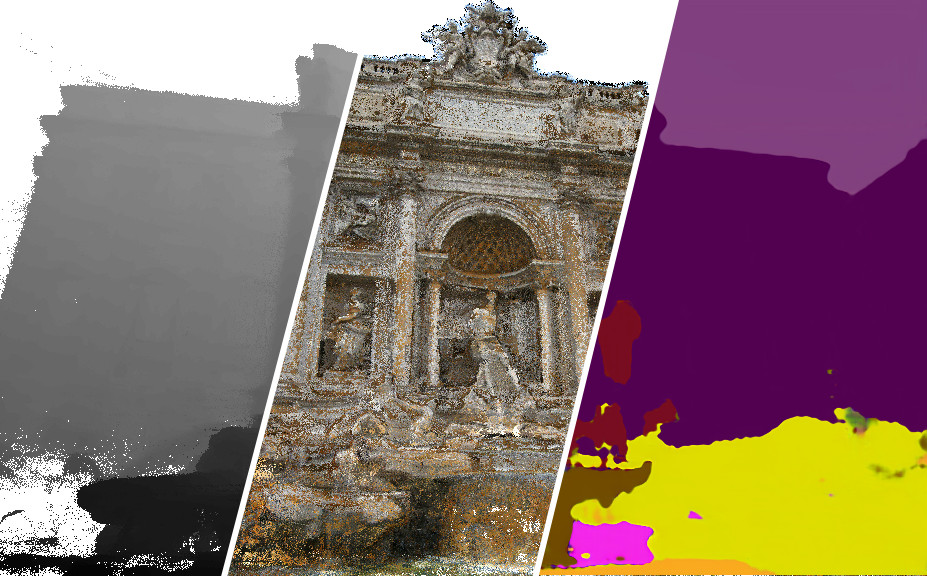}%
        \vspace*{.03in}  %
        \caption{Input deep buffer}
    \end{subfigure} 
    \begin{subfigure}[b]{0.49\linewidth}
        \includegraphics[width=\linewidth]{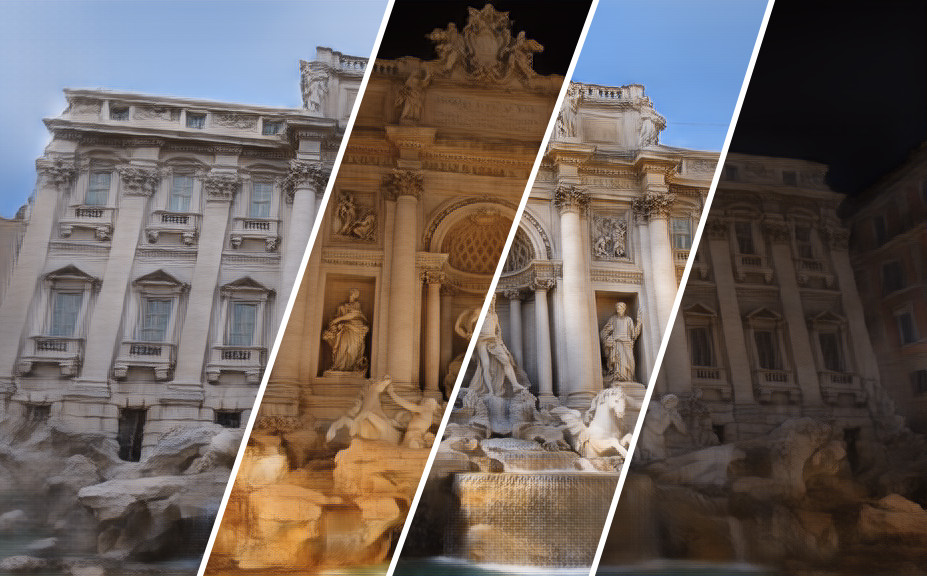}%
        \vspace*{.03in}  %
        \caption{Output renderings}
    \end{subfigure}%
    \caption{\emph{Neural Rerendering in the Wild}~\cite{Meshry_2019_CVPR} reconstructs a proxy 3D model from a large-scale internet photo collection. This model is rendered into a deep buffer of depth, color and semantic labels (a). A neural rerendering network translates these buffers into realistic renderings under multiple appearances (b). \newstuff{\footnotesize{Images taken from Meshry et al.~\cite{Meshry_2019_CVPR}}}.
	}
    \label{fig:neural-wild}
    \vspace{-0.2cm}
\end{figure}

%% file: applications/mpi_based.tex
\subsubsection{Novel View Synthesis with Multiplane Images}
\label{subsec:deepviewsynthesis}

Given a sparse set of input views of an object, Xu et al. \cite{xu2019deepviewsynthesis} also address the problem of rendering the object from novel viewpoints (see \Cref{fig:deep_view_synthesis}).
Unlike previous view interpolation methods that work with images captured under natural illumination and at a small baseline, they aim to capture the light transport of the scene, including view-dependent effects like specularities.
Moreover they attempt to do this from a sparse set of images captured at large baselines, in order to make the capture process more light-weight.
They capture six images of the scene under point illumination in a cone of about $60^{\circ}$ and render any novel viewpoint within this cone.
The input images are used to construct a plane sweeping volume aligned with the novel viewpoint\cite{flynn2016deepstereo}.
This volume is processed by 3D CNNs to reconstruct both scene depth and appearance.
To handle the occlusions caused by the large baseline, they propose predicting attention maps that capture the visibility of the input viewpoints at different pixels.
These attention maps are used to modulate the appearance plane sweep volume and remove inconsistent content.
The network is trained on synthetically rendered data with supervision on both geometry and appearance; at test time it is able to synthesize photo-realistic results of real scenes featuring high-frequency light transport effects such as shadows and specularities.   
\input{applications/figures/deep_view_synthesis.tex}
\textit{DeepView}~\cite{Flynn19DeepView} is a technique to visualize light fields under novel views.
The view synthesis is based on multi-plane images~\cite{zhou2018stereomag} that are estimated by a learned gradient descent method given a sparse set of input views.
Similar to image-based rendering, the image planes can be warped to new views and are rendered back-to-front into the target image.

%% file: applications/figures/deep_view_synthesis.tex
\begin{figure}[t]
	\centering
	\includegraphics[width=0.85\linewidth]{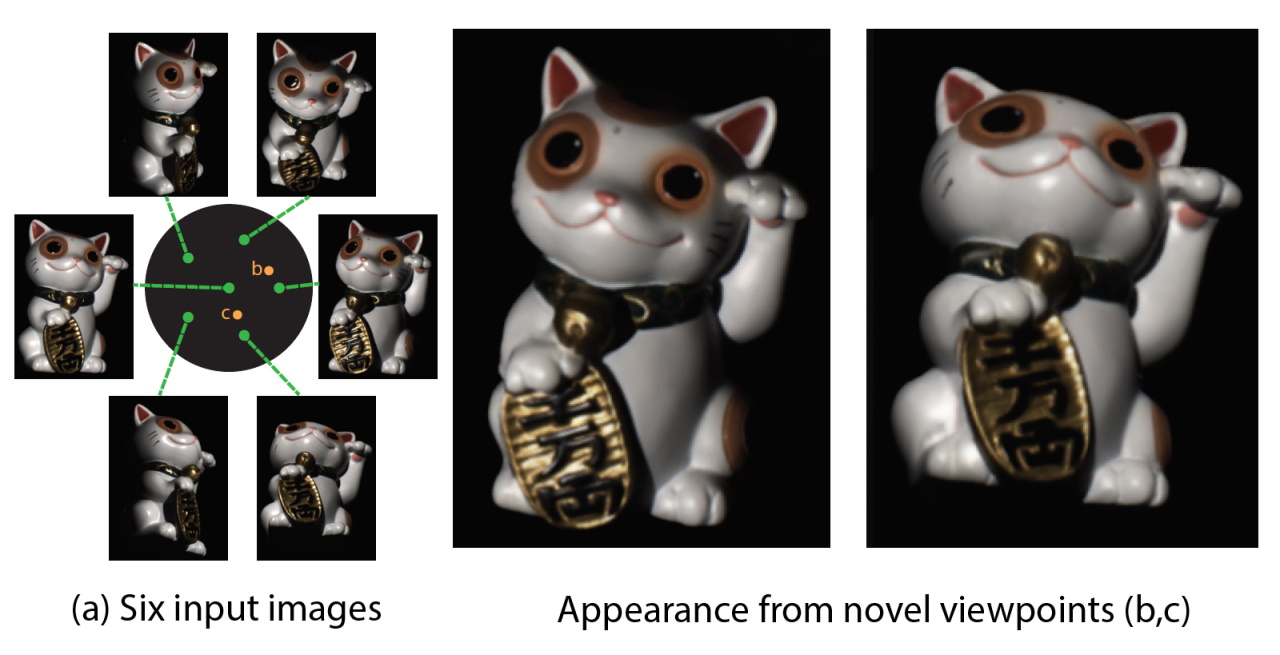}
	\caption
	{
	   Xu et al. \cite{xu2019deepviewsynthesis} render scene appearance from a novel viewpoint, given only six sparse, wide baseline views. \newstuff{\footnotesize{Images taken from Xu~\etal~\cite{xu2019deepviewsynthesis}}}.
	}
	\label{fig:deep_view_synthesis}
	\vspace{-0.5cm}
\end{figure}

%% file: applications/neural_scene_reps_and_rendering.tex
\subsubsection{Neural Scene Representation and Rendering}
\label{app:nsrr}

While neural rendering methods based on multi-plane images and image-based rendering have enabled some impressive results, they prescribe the model's internal representation of the scene as a point cloud, a multi-plane image, or a mesh, and do not allow the model to learn an optimal representation of the scene's geometry and appearance. A recent line in novel view synthesis is thus to build models with neural scene representations: learned, feature-based representations of scene properties.
The Generative Query Network~\cite{eslami2018neural} is a framework for learning a low-dimensional feature embedding of a scene, explicitly modeling the stochastic nature of such a neural scene representation due to incomplete observations. A scene is represented by a collection of observations, where each observation is a tuple of an image and its respective camera pose. Conditioned on a set of context observations and a target camera pose, the GQN parameterizes a distribution over frames observed at the target camera pose, consistent with the context observations. The GQN is trained by maximizing the log-likelihood of each observation given other observations of the same scene as context.
Given several context observations of a single scene, a convolutional encoder encodes each of them into a low-dimensional latent vector. These latent vectors are aggregated to a single representation $r$ via a sum. A convolutional Long-Short Term Memory network (ConvLSTM) parameterizes an auto-regressive prior distribution over latent variables $z$. At every timestep, the hidden state of the ConvLSTM is decoded into a residual update to a canvas $u$ that represents the sampled observation. To make the optimization problem tractable, the GQN uses an approximate posterior at training time.
The authors demonstrate the capability of the GQN to learn a rich feature representation of the scene on novel view synthesis, control of a simulated robotic arm, and the exploration of a labyrinth environment. The probabilistic formulation of the GQN allows the model to sample different frames all consistent with context observations, capturing, for instance, the uncertainty about parts of the scene that were occluded in context observations.

%% file: applications/voxel_based.tex
\subsubsection{Voxel-based Novel View Synthesis Methods}
\label{app:rendernet}

While learned, unstructured neural scene representations are an attractive alternative to hand-crafted scene representations, they come with a number of drawbacks. First and foremost, they disregard the natural 3D structure of scenes. As a result, they fail to discover multi-view and perspective geometry in regimes of limited training data. 
\newstuff{Inspired by recent progress in geometric deep learning~\cite{kar2017stereo,choy20163d,Rezende2016unsupervised,henzler2018escaping}, a line of neural rendering approaches has emerged that instead proposes to represent the scene as a voxel grid, thus enforcing 3D structure.}

\textit{RenderNet}~\cite{nguyen2018rendernet} proposes a convolutional neural network architecture that implements differentiable rendering from a scene explicitly represented as a 3D voxel grid. The model is retrained for each class of objects and requires tuples of images with labeled camera pose. RenderNet enables novel view synthesis, texture editing, relighting, and shading. Using the camera pose, the voxel grid is first transformed to camera coordinates. A set of 3D convolutions extracts 3D features. The 3D voxel grid of features is translated to a 2D feature map via a subnetwork called the ``projection unit.'' The projection unit first collapses the final two channels of the 3D feature voxel grid and subsequently reduces the number of channels via 1x1 2D convolutions. The 1x1 convolutions have access to all features along a single camera ray, enabling them to perform projection and visibility computations of a typical classical renderer. Finally, a 2D up-convolutional neural network upsamples the 2D feature map and computes the final output.
The authors demonstrate that RenderNet learns to render high-resolution images from low-resolution voxel grids. RenderNet can further learn to apply varying textures and shaders, enabling scene relighting and novel view synthesis of the manipulated scene. They further demonstrate that RenderNet may be used to recover a 3D voxel grid representation of a scene from single images via an iterative reconstruction algorithm, enabling subsequent manipulation of the representation.

\label{app:deepvoxels}
\emph{DeepVoxels}~\cite{sitzmann2019deepvoxels} enables joint reconstruction of geometry and appearance of a scene and subsequent novel view synthesis. DeepVoxels is trained on a specific scene, given only images as well as their extrinsic and intrinsic camera parameters -- no explicit scene geometry is required. This is achieved by representing a scene as a Cartesian 3D grid of embedded features, combined with a network architecture that explicitly implements image formation using multi-view and projective geometry operators. Features are first extracted from 2D observations. 2D features are then un-projected by replicating them along their respective camera rays, and integrated into the voxel grid by a small 3D U-net. To render the scene with given camera extrinsic and intrinsic parameters, a virtual camera is positioned in world coordinates. Using the intrinsic camera parameters, the voxel grid is resampled into a canonical view volume. To reason about occlusions, the authors propose an occlusion reasoning module. The occlusion module is implemented as a 3D U-Net that receives as input all the features along a camera ray as well as their depth, and produces as output a visibility score for each feature along the ray, where the scores along each ray sum to one. The final projected feature is then computed as a weighted sum of features along each ray. Finally, the resulting 2D feature map is translated to an image using a small U-Net. As a side-effect of the occlusion reasoning module, DeepVoxels produces depth maps in an unsupervised fashion. The model is fully differentiable from end to end, and is supervised only by a 2D re-rendering loss enforced over the training set. The paper shows wide-baseline novel view synthesis on several challenging scenes, both synthetic and real, and outperforms baselines that do not use 3D structure by a wide margin.

\textit{Visual Object Networks (VONs)}~\cite{zhu2018visual} is a 3D-aware generative model for synthesizing the appearance of objects with a disentangled 3D representation. Inspired by classic rendering pipelines, VON decomposes the neural image formation model into three factors---viewpoint, shape, and texture. The model is trained with an end-to-end adversarial learning framework that jointly learns the distribution of 2D images and 3D shapes through a differentiable projection module. During test time, VON can synthesize a 3D shape, its intermediate 2.5D depth representation, and a final 2D image all at once. This 3D disentanglement allows users to manipulate the shape, viewpoint, and texture of an object independently.

\textit{HoloGAN}~\cite{hologan} builds on top of the learned projection unit of RenderNet to build an unconditional generative model that allows explicit viewpoint changes. 
It implements an explicit affine transformation layer that directly applies view manipulations to learnt 3D features. 
As in DeepVoxels, the network learns a 3D feature space, but more bias about the 3D object / scene is introduced by transforming these deep voxels with a random latent vector $z$.
In this way, an unconditional GAN that natively supports viewpoint changes can be trained in an unsupervised fashion.
Notably, HoloGAN requires neither pose labels and intrinsic camera information nor multiple views of an object.

%% file: applications/implicit_based.tex
\subsubsection{Implicit-function based Approaches}
While 3D voxel grids have demonstrated that a 3D-structured scene representation benefits multi-view consistent scene modeling, their memory requirement scales cubically with spatial resolution, and they do not parameterize surfaces smoothly, requiring a neural network to learn priors on shape as joint probabilities of neighboring voxels. 
As a result, they cannot parameterize large scenes at a sufficient spatial resolution, and have so far failed to generalize shape and appearance across scenes, which would allow applications such as reconstruction of scene geometry from only few observations.
In geometric deep learning, recent work alleviated these problems by modeling geometry as the level set of a neural network~\cite{park2019deepsdf, mescheder2019occupancy}. Recent neural rendering work generalizes these approaches to allow rendering of full color images.
\label{app:pifu}
In addition to parameterizing surface geometry via an implicit function, \emph{Pixel-Aligned Implicit Functions}~\cite{saito2019pifu} represent object color via an implicit function. An image is first encoded into a pixel-wise feature map via a convolutional neural network. A fully connected neural network then takes as input the feature at a specific pixel location as well as a depth value $z$, and classifies the depth as inside/outside the object. The same architecture is used to encode color. The model is trained end-to-end, supervised with images and 3D geometry. The authors demonstrate single- and multi-shot 3D reconstruction and novel view synthesis of clothed humans.
\label{app:srns}
\input{applications/figures/srns.tex}
\emph{Scene Representation Networks} (SRNs) ~\cite{sitzmann2019srns} encodes both scene geometry and appearance in a single fully connected neural network, the SRN, that maps world coordinates to a feature representation of local scene properties. A differentiable, learned neural ray-marcher is trained end-to-end given only images and their extrinsic and intrinsic camera parameters---no ground-truth shape information is required.
The SRN takes as input $(x, y, z)$ world coordinates and computes a feature embedding. To render an image, camera rays are traced to their intersections with scene geometry (if any) via a differentiable, learned raymarcher, which computes the length of the next step based on the feature returned by the SRN at the current intersection estimate. The SRN is then sampled at ray intersections, yielding a feature for every pixel. This 2D feature map is translated to an image by a per-pixel fully connected network. Similarly to DeepSDF~\cite{park2019deepsdf}, SRNs generalize across scenes in the same class by representing each scene by a code vector $\mathbf{z}$. The code vectors $\mathbf{z}$ %
are mapped to the parameters of a SRN via a fully connected neural network, a so-called hypernetwork. The parameters of the hypernetwork are jointly optimized with the code vectors and the parameters of the pixel generator.
The authors demonstrate single-image reconstruction of geometry and appearance of objects in the ShapeNet dataset (\Cref{fig:srns}), as well as multi-view consistent view synthesis. Due to their per-pixel formulation, SRNs generalize to completely unseen camera poses like zoom or camera roll.

%% file: applications/figures/srns.tex
\begin{figure}[t]
     \centering
         \includegraphics[width=\linewidth]{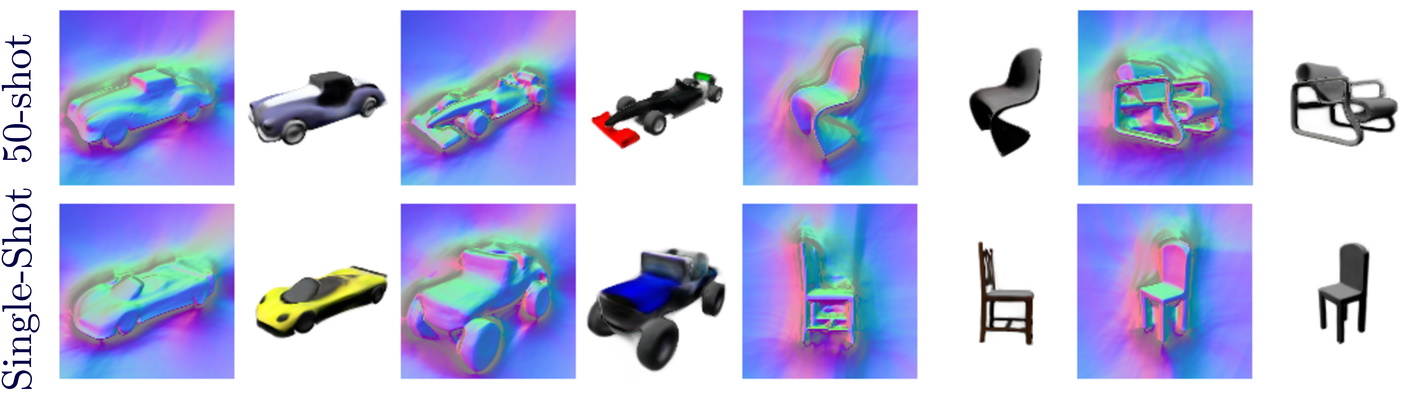}
         \caption{\emph{Scene Representation Networks}~\cite{sitzmann2019srns} allow reconstruction of scene geometry and appearance from images via a continuous, 3D-structure-aware neural scene representation, and subsequent, multi-view-consistent view synthesis. By learning strong priors, they allow full 3D reconstruction from only a single image (bottom row, surface normals and color render). \newstuff{\footnotesize{Images taken from Sitzmann et al.~\cite{sitzmann2019srns}}}.
         }
\label{fig:srns}
\vspace{-0.2cm}
\end{figure}

%% file: applications/free_viewpoint_videos.tex
\vspace{-0.1cm}
\subsection{Free Viewpoint Videos}
\label{subsec:free_viewpoint}

\textit{Free Viewpoint Videos}, also known as \textit{Volumetric Performance Capture}, rely on multi-camera setups to acquire the 3D shape and texture of performers. The topic has gained a lot of interest in the research community starting from the early work of Tanco and Hilton \cite{hilton2} and reached compelling high quality results with the works\ of Collet et al. \cite{fvv} and its real-time counterpart by Dou et al. \cite{dou16,dou17}. 
Despite the efforts, these systems lack photorealism due to missing high frequency details \cite{holoportation} or baked in texture \cite{fvv}, which does not allow for accurate and convincing re-lighting of these models in arbitrary scenes. Indeed, volumetric performance capture methods lack view dependent effects (\eg specular highlights); moreover, imperfections in the estimated geometry usually lead to blurred texture maps. Finally, creating temporally consistent 3D models \cite{fvv} is very challenging in many real world cases (e.g. hair, translucent materials).
A recent work on human performance capture by Guo~\etal~\cite{relightables} overcomes many of these limitations by combining traditional image based relighting methods \cite{lightstage1} with recent advances in high-speed and accurate depth sensing \cite{need4speed,sos}.
In particular, this system uses $58$ $12.4$MP RGB cameras combined with $32$ $12.4$MP active IR sensors to recover very accurate geometry.
During the capture, the system interleaves two different lighting conditions based on spherical gradient illumination \cite{Fyffe:2009}. This produces an unprecedented level of photorealism for a volumetric capture pipeline (\Cref{fig:relightables}).
\input{applications/figures/relightables.tex}
Despite steady progress and encouraging results obtained by these 3D capture systems, they still face important challenges and limitations. Translucent and transparent objects cannot be easily captured; reconstructing thin structures (e.g. hair) is still very challenging even with high resolution depth sensors. Nevertheless, these multi-view setups provide the foundation for machine learning methods \cite{Martin-Brualla:2018:LEP:3272127.3275099,Lombardi:2018:DAM:3197517.3201401,Lombardi:2019:NVL:3306346.3323020,Pandey_2019_CVPR}, which heavily rely on training data to synthesize high quality humans in arbitrary views and poses.
\subsubsection{LookinGood with Neural Rerendering}

The \emph{LookinGood} system by Martin-Brualla et al. \cite{Martin-Brualla:2018:LEP:3272127.3275099} introduced the concept of neural rerendering for performance capture of human actors.
The framework relies on a volumetric performance capture system \cite{dou17}, which reconstructs the performer in real-time. These models can then be rendered from arbitrary viewpoints using the known geometry. Due to real-time constraints, the reconstruction quality suffers from many artifacts such as missing depth, low resolution texture and oversmooth geometry. Martin-Brualla et al. propose to add ``witness cameras'', which are high resolution RGB sensors ($12$MP) that are not used in the capture system, but can provide training data for a deep learning architecture to re-render the output of the geometrical pipeline. The authors show that this enables high quality re-rendered results for arbitrary viewpoints, poses and subjects in real-time.
\input{applications/figures/lookingood.tex}
The problem at hand is very interesting since it is tackling denoising, in-painting and super-resolution simultaneously and in real-time. In order to solve this, authors cast the task into an image-to-image translation problem \cite{pix2pix2016} and they introduce a semantic aware loss function that uses the semantic information (available at training time) to increase the weights of the pixels belonging to salient areas of the image and to retrieve precise silhouettes ignoring the contribution of the background. The overall system is shown in \Cref{fig:lookingood}.

\input{applications/neural_volumes.tex}

\subsubsection{Free Viewpoint Videos from a Single Sensor}

The availability of multi-view images at training and test time is one of the key elements for the success of free viewpoint systems. However this capture technology is still far from being accessible to a typical consumer who, at best, may own a single RGBD sensor such as a Kinect. Therefore, parallel efforts \cite{Pandey_2019_CVPR} try to make the capture technology accessible through consumer hardware by dropping the infrastructure requirements through deep learning. 
Reconstructing performers from a single image is very related to the topic of synthesizing humans in unseen poses \cite{acm18,balakrishnan_cvpr18,si_2018_CVPR,ma18,ma17,Guler2018DensePose,Chan2018}. Differently from the other approaches, the recent work of Pandey et al. \cite{Pandey_2019_CVPR} synthesizes performers in  \textit{unseen poses} and from \textit{arbitrary viewpoints}, mimicking the behavior of volumetric capture systems. The task at hand is much more challenging because it requires disentangling pose, texture, background and viewpoint.
Pandey et al. propose to solve this problem by leveraging a semi-parametric model. In particular they assume that a short calibration sequence of the user is available: e.g. the user rotates in front of the camera before the system starts. Multiple deep learning stages learn to combine the current viewpoint (which contains the correct user pose and expression) with the pre-recorded calibration images (which contain the correct viewpoint but wrong poses and expressions). The results are compelling given the substantial reduction in the infrastructure required.%

%% file: applications/figures/relightables.tex
\begin{figure}[t]
	\centering
	\includegraphics[width=\linewidth]{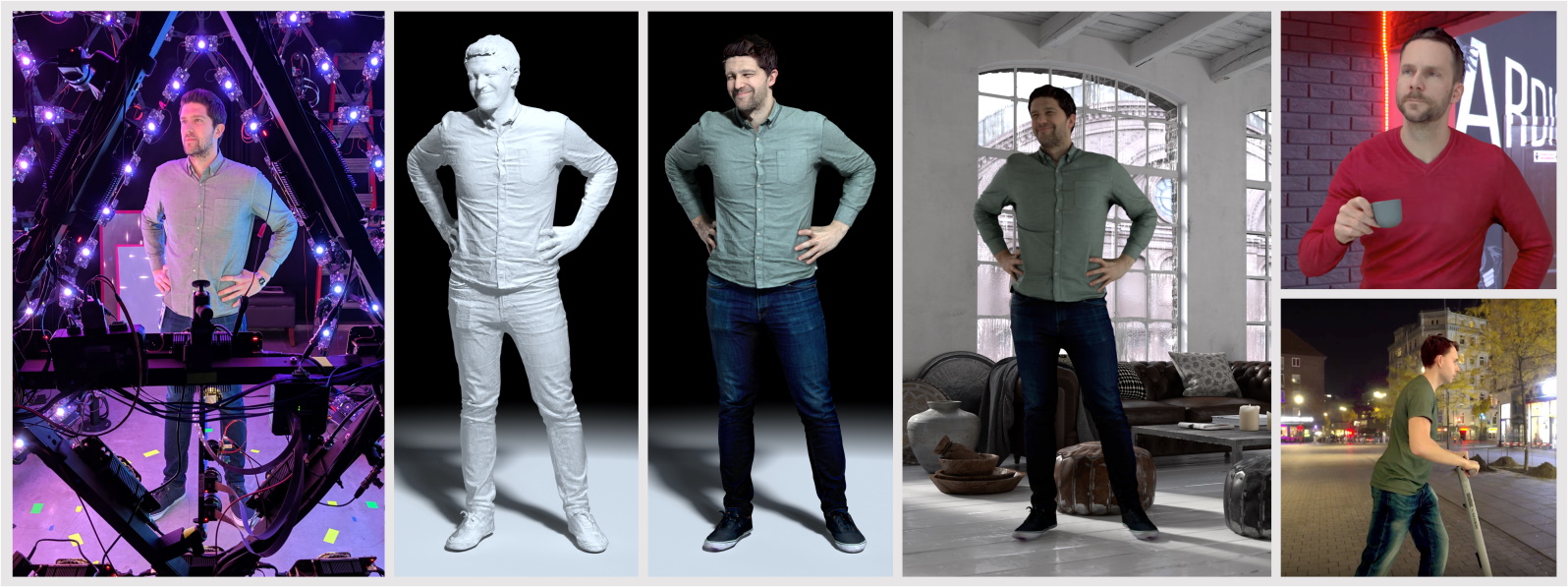}
	\caption
	{
	   \emph{The Relightables} system by Guo et al. \cite{relightables} for free viewpoint capture of humans with realistic re-lighting. From left to right: a performer captured in the Lightstage, the estimated geometry and albedo maps, examples of relightable volumetric videos. 
	   \newstuff{\footnotesize{Images taken from Guo et al.~\cite{relightables}}}.
	}
	\vspace{-0.5cm}
	\label{fig:relightables}
\end{figure}

%% file: applications/figures/lookingood.tex
\begin{figure}[t]
	\centering
	\includegraphics[width=\linewidth]{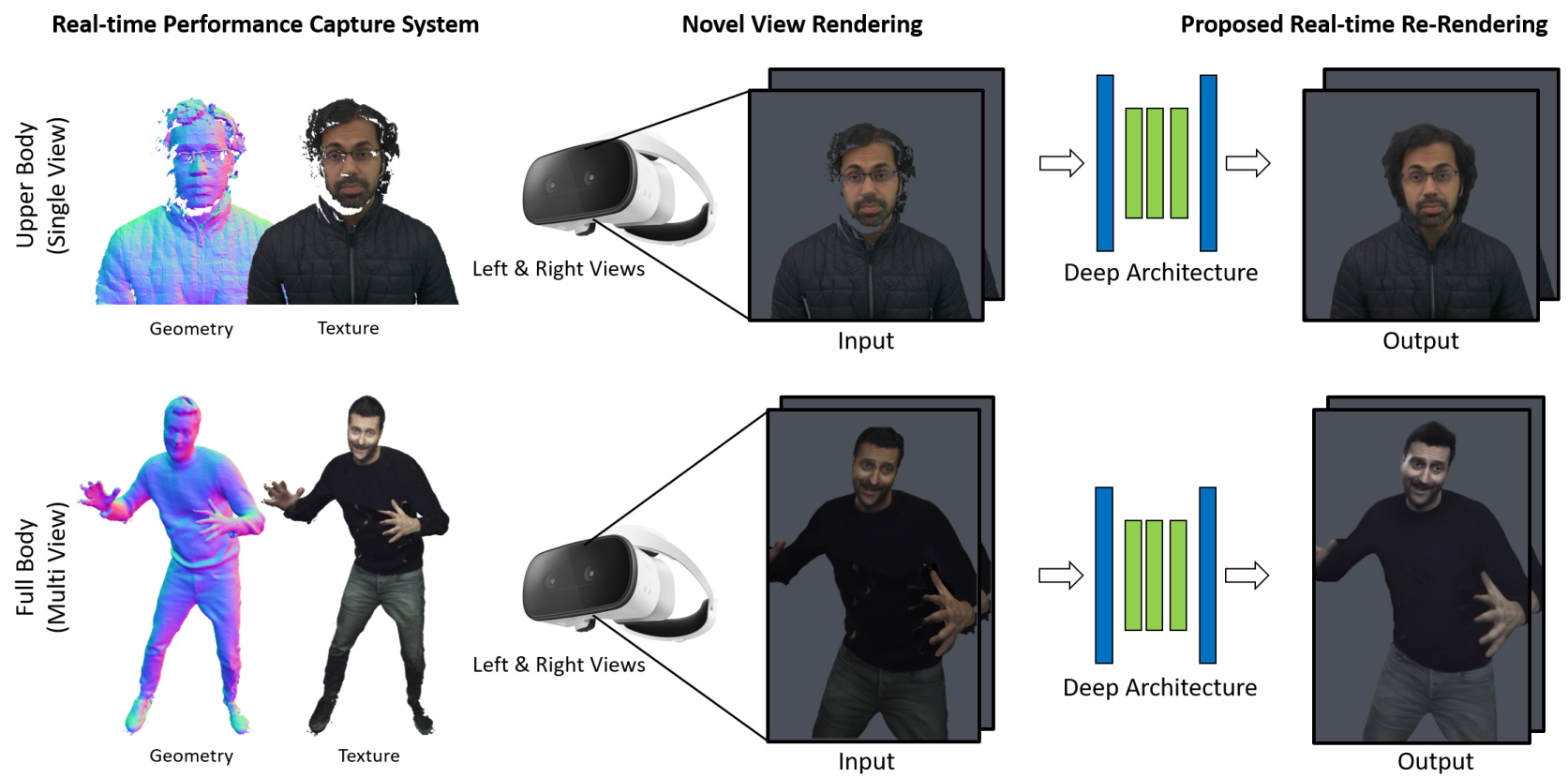}
	\caption
	{
	   The \emph{LookinGood} system \cite{Martin-Brualla:2018:LEP:3272127.3275099} uses real-time neural re-rendering to enhance performance capture systems. 
	   \newstuff{\footnotesize{Images taken from Martin-Brualla et al.~\cite{Martin-Brualla:2018:LEP:3272127.3275099}}}.
	}
	\label{fig:lookingood}
\end{figure}

%% file: applications/neural_volumes.tex
\subsubsection{Neural Volumes}
\label{app:neural_volumes}

\emph{Neural Volumes} \cite{Lombardi:2019:NVL:3306346.3323020} addresses the problem of automatically creating, rendering, and animating high-quality object models from multi-view video data (see \Cref{fig:neuralvolumes}). The method trains a neural network to encode frames of a multi-view video sequence into a compact latent code which is decoded into a semi-transparent volume containing RGB and opacity values at each $(x,y,z)$ location. The volume is rendered by raymarching from the camera through the volume, accumulating color and opacity to form an output image and alpha matte. Formulating the problem in 3D rather than in screen space has several benefits: viewpoint interpolation is improved because the object must be representable as a 3D shape, and the method can be easily combined with traditional triangle-mesh rendering.
The method produces high-quality models despite using a low-resolution voxel grid ($128^3$) by introducing a learned warp field that not only helps to model the motion of the scene 
but also reduces blocky voxel grid artifacts by deforming voxels to better match the geometry of the scene and allows the system to shift voxels to make better use of the voxel resolution available.
The warp field is modeled as a spatially-weighted mixture of affine warp fields, which can naturally model piecewise deformations.
By virtue of the semi-transparent volumetric representation, the method can reconstruct challenging objects such as moving hair, fuzzy toys, and smoke all from only 2D multi-view video with no explicit tracking required. The latent space encoding enables animation by generating new latent space trajectories or by conditioning the decoder on some information like head pose.

\input{applications/figures/neural_volumes.tex}

%% file: applications/figures/neural_volumes.tex
\begin{figure}[t]
	\centering
	\includegraphics[width=\linewidth]{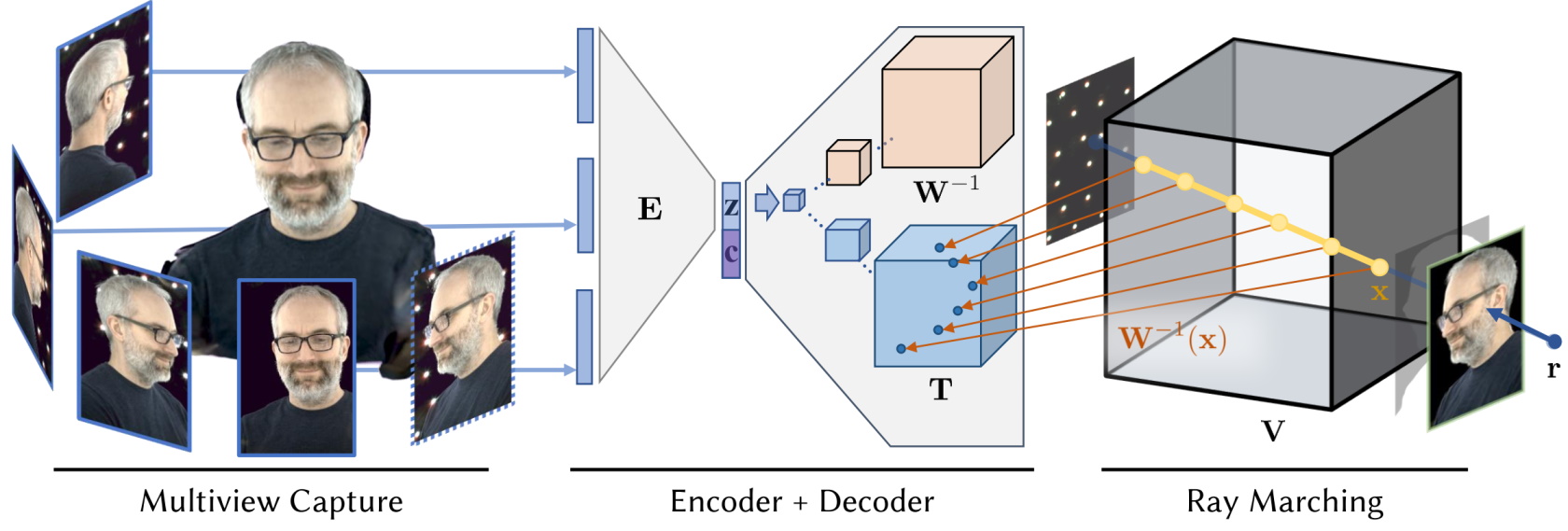}
	\vspace{-0.5cm}
	\caption
	{
	   Pipeline for \emph{Neural Volumes} \cite{Lombardi:2019:NVL:3306346.3323020}.
	   Multi-view capture is input to an encoder to produce a latent code $\mathbf{z}$. $\mathbf{z}$ is decoded to a volume that stores RGB$\alpha$ values, as well as a warp field. Differentiable ray marching renders the volume into an image, allowing the system to be trained by minimizing the difference between rendered and target images. \newstuff{\footnotesize{Images taken from Lombardi et al.~\cite{Lombardi:2019:NVL:3306346.3323020}}}.
	}
	\label{fig:neuralvolumes}
	\vspace{-0.5cm}
\end{figure}

%% file: applications/relighting.tex
\subsection{Learning to Relight}
\label{subsec:relighting}

Photo-realistically rendering of a scene under novel illumination---a procedure known as ``relighting''---is a fundamental component of a number of graphics applications including compositing, augmented reality and visual effects. An effective way to accomplish this task is to use image-based relighting methods that take as input images of the scene captured under different lighting conditions (also known as a ``reflectance field''), and combine them to render the scene's appearance under novel illumination~\cite{lightstage1}. Image-based relighting can produce high-quality, photo-realistic results and has even been used for visual effects in Hollywood productions. However, these methods require slow data acquisition with expensive, custom hardware, precluding the applicability of such methods to settings like dynamic performance and ``in-the-wild'' capture.
Recent methods address these limitations by using synthetically rendered or real, captured reflectance field data to train deep neural networks that can relight scenes from just a few images.
\subsubsection{Deep Image-based Relighting from Sparse Samples}

Xu et al.\cite{xu2018deeprelighting} propose an image-based relighting method that can relight a scene from a sparse set of five images captured under learned, optimal light directions. Their method uses a deep convolutional neural network to regress a relit image under an arbitrary directional light from these five images. Traditional image-based relighting methods rely on the \emph{linear} superposition property of lighting, and thus require tens to hundreds of images for high-quality results. Instead, by training a \emph{non-linear} neural relighting network, this method is able to accomplish relighting from sparse images. The relighting quality depends on the input light directions, and the authors propose combining a custom-designed sampling network with the relighting network, in an end-to-end fashion, to jointly learn both the optimal input light directions and the relighting function. The entire system is trained on a large synthetic dataset comprised of procedurally generated shapes rendered with complex, spatially-varying reflectances. At test time, the method is able to relight real scenes and reproduce complex, high-frequency lighting effects like specularities and cast shadows. 

\input{applications/figures/deep_relighting.tex}

\subsubsection{Multi-view Scene Relighting}

Given multiple views of a large-scale outdoor scene captured under uncontrolled natural illumination, Philip et al.\cite{philip2019multiviewrelighting} can render the scene under novel outdoor lighting (parameterized by the sun position and cloudiness level).%
The input views are used to reconstruct the 3D geometry of the scene; this geometry is coarse and erroneous and directly relighting it would produce poor results. Instead, the authors propose using this geometry to construct intermediate buffers---normals, reflection features, and RGB shadow maps---as auxiliary inputs to guide a neural network-based relighting method. The method also uses a shadow refinement network to improve the removal and addition of shadows that are an important cue in outdoor images. While the entire method is trained on a synthetically rendered dataset, it generalizes to real scenes, producing high-quality results for applications like the creation of time-lapse effects from multiple (or single) images and relighting scenes in traditional image-based rendering pipelines.

\subsubsection{Deep Reflectance Fields}
\emph{Deep Reflectance Fields} \cite{meka2019deep} presents a novel technique to relight images of human faces by learning a model of facial reflectance from a database of 4D reflectance field data of several subjects in a variety of expressions and viewpoints. Using a learned model, a face can be relit in arbitrary illumination environments using only two original images recorded under spherical color gradient illumination \cite{Fyffe:2009}. The high-quality results of the method indicate that the color gradient images contain the information needed to estimate the full 4D reflectance field, including specular reflections and high frequency details. While capturing images under spherical color gradient illumination still requires a special lighting setup, reducing the capture requirements to just two illumination conditions, as compared to previous methods that require hundreds of images \cite{lightstage1}, allows the technique to be applied to \textit{dynamic} facial performance capture (\Cref{fig:deep_reflectance}).

\input{applications/figures/deep_reflectance_fields.tex}

\subsubsection{Single Image Portrait Relighting}

A particularly useful application of relighting methods is to change the lighting of a portrait image captured in the wild, i.e., with off-the-shelf (possibly cellphone) cameras under natural unconstrained lighting. While the input in this case is only a single (uncontrolled) image, recent methods have demonstrated state-of-the-art results using deep neural networks \cite{sun2019single,zhou2019portraitrelighting}. The relighting model in these methods consists of a deep neural network that has been trained to take a single RGB image as input and produce as output a relit version of the portrait image under an arbitrary user-specified environment map. Additionally, the model also predicts an estimation of the current lighting conditions and, in the case of Sun et al.\cite{sun2019single}, can run on mobile devices in \textasciitilde160ms, see \Cref{fig:portrait_relighting}.
Sun et al.~represent the target illumination as an environment map and train their network using captured reflectance field data. On the other hand, Zhou et al. \cite{zhou2019portraitrelighting} use a spherical harmonics representation for the target lighting and train the network with a synthetic dataset created by relighting single portrait images using a traditional ratio image-based method.
Instead of having an explicit inverse rendering step for estimating geometry and reflectance \cite{barron2014shape, shu2017neural, sengupta2018sfsnet}, these methods directly regress to the final relit image from an input image and a ``target'' illumination. In doing so, they bypass restrictive assumptions like Lambertian reflectance and low-dimensional shape spaces that are made in traditional face relighting methods, and are able to generalize to full portrait image relighting including hair and accessories. 

\input{applications/figures/portrait_relighting.tex}

%% file: applications/figures/deep_relighting.tex
\begin{figure}[t]
	\centering
	\includegraphics[width=\linewidth]{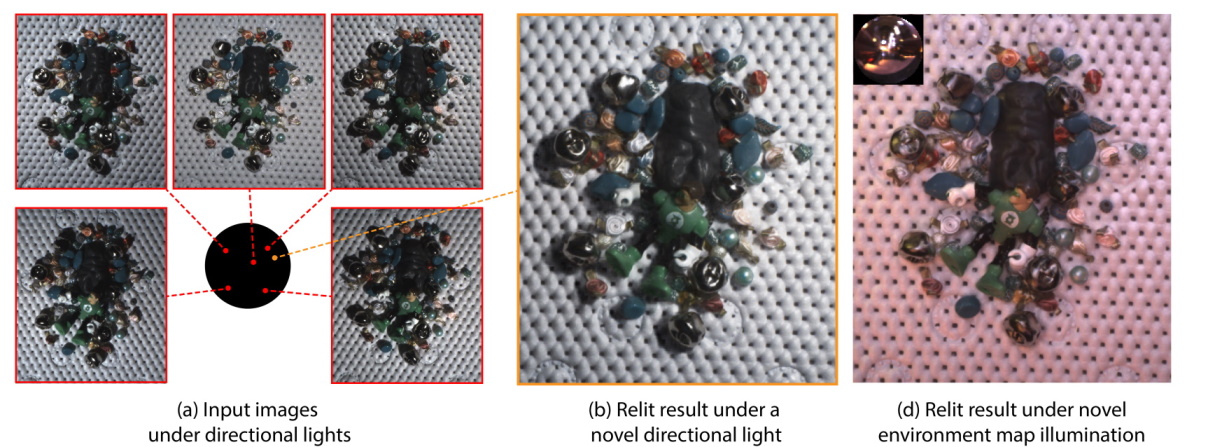}
	\caption
	{
	   Xu et al. \cite{xu2018deeprelighting} are able to generate relit versions of a scene under novel directional and environment map illumination from only five images captured under specified directional lights. \newstuff{\footnotesize{Images taken from Xu~\etal~\cite{xu2018deeprelighting}}}.
	}
	\label{fig:deep_relighting}
	\vspace{-0.2cm}
\end{figure}

%% file: applications/figures/deep_reflectance_fields.tex
\begin{figure}[t]
	\centering
	\includegraphics[width=\linewidth]{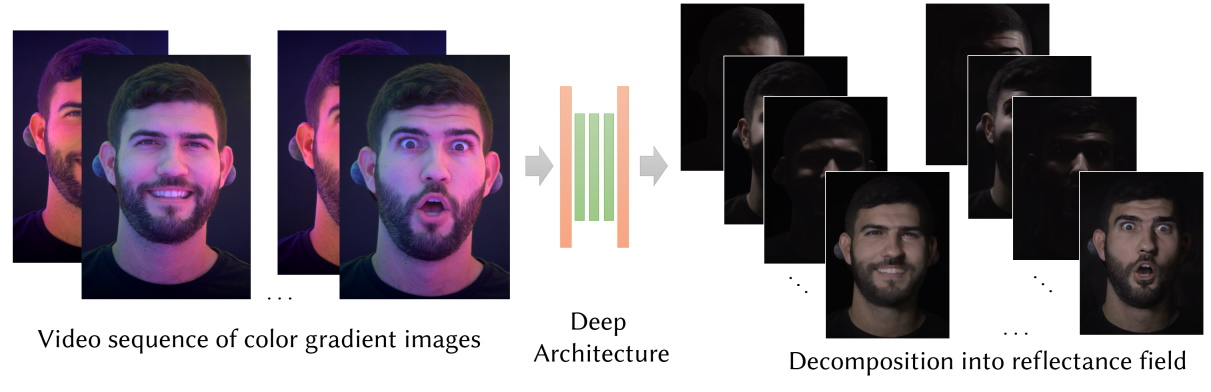}
	\vspace{-0.5cm}
	\caption
	{
	   Meka et al. \cite{meka2019deep} decompose the full reflectance fields by training a convolutional neural network that maps two spherical gradient images to any one-light-at-a-time image. 
	   \newstuff{\footnotesize{Images taken from Meka et al.~\cite{meka2019deep}}}.
	}
	\label{fig:deep_reflectance}
	\vspace{-0.1cm}
\end{figure}

%% file: applications/figures/portrait_relighting.tex
\begin{figure}[t]
	\centering
	\includegraphics[width=\linewidth]{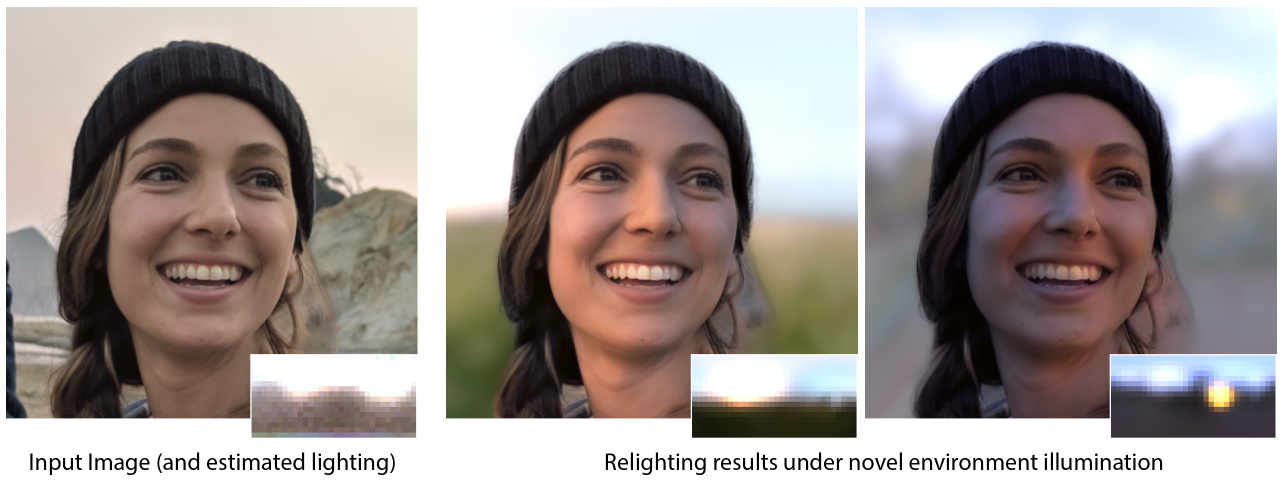}
	\vspace{-0.4cm}
	\caption
	{
	    Given a single portrait image captured with a standard camera, portrait relighting methods \cite{sun2019single} can generate images of the subject under novel lighting environments. 
	    \newstuff{\footnotesize{Images taken from Sun et al.~\cite{sun2019single}}}.
	}
	\label{fig:portrait_relighting}
	\vspace{-0.4cm}
\end{figure}

%% file: applications/facial_reenactment.tex
\subsection{Facial Reenactment}
\label{subsec:face_body}
Facial reenactment aims to modify scene properties beyond those of viewpoint (\Cref{subsec:free_viewpoint}) and lighting (\Cref{subsec:relighting}), for example by generating new head pose motion, facial expressions, or speech.  Early methods were based on classical computer graphics techniques. 
While some of these approaches only allow implicit control, i.e., retargeting facial expressions from a source to a target sequence~\cite{thies2016face}, explicit control has also been explored~\cite{blanz2003reanimating}.
These approaches usually involve reconstruction of a 3D face model from the input, followed by editing and rendering of the model to synthesize the edited result. 
Neural rendering techniques overcome the limitations of classical approaches by better dealing with inaccurate 3D reconstruction and tracking, as well as better photorealistic appearance rendering.  
Early neural rendering approaches, such as that of Kim~\etal~\cite{Kim:2018:DVP:3197517.3201283}, use a conditional GAN to refine the outputs estimated by classical methods.
In addition to more photo-realistic results compared to classical techniques, neural rendering methods allow for the control of head pose in addition to facial expressions~\cite{Kim:2018:DVP:3197517.3201283,zakharov2019few,nagano2018pagan,Wiles18}. 
Most neural rendering approaches for facial reenactment are trained separately for each identity. 
Only recently, methods which generalize over multiple identities have been explored~\cite{zakharov2019few,Wiles18,nagano2018pagan,Geng:2018:WGS:3272127.3275043}.

\input{applications/dvp.tex}

\input{applications/text-based-editing.tex}

\input{applications/neural-textures.tex}

\input{applications/talking_head_models.tex}
\input{applications/deep_appearance_models.tex}
While neural rendering approaches for facial reenactment achieve impressive results, many challenges still remain to be solved. 
Full head reenactment, including control over the head pose, is very challenging in dynamic environments. 
Many of the methods discussed do not preserve high-frequency details in a temporally coherent manner. 
In addition, photorealistic synthesis and editing of hair motion and  mouth interior including tongue motion is challenging. 
Generalization across different identities without any degradation in quality is still an open problem.

%% file: applications/dvp.tex
\subsubsection{Deep Video Portraits}
\label{app:dvp}
\emph{Deep Video Portraits} \cite{Kim:2018:DVP:3197517.3201283} is a system for full head reenactment of portrait videos. 
The head pose, facial expressions and eye motions of the person in a video are transferred from another reference video. 
A facial performance capture method is used to compute 3D face reconstructions for both reference and target videos.
This reconstruction is represented using a low-dimensional semantic representation  which includes identity, expression, pose, eye motion, and illumination parameters.
Then, a rendering-to-video translation network, based on U-Nets, is trained to convert classical computer graphics renderings of the 3D models to photo-realistic images. 
The network adds photo-realistic details on top of the imperfect face renderings, in addition to completing the scene by adding hair, body, and background. 
The training data consists of pairs of training frames, and their corresponding 3D reconstructions.
Training is identity and scene specific, with only a few minutes (typically 5-10) of training data needed. 
At test time, semantic dimensions which are relevant for reenactment, i.e., expressions, eye motion and rigid pose are transferred from a source to a different target 3D model. 
The translation network subsequently converts the new 3D sequence into a photo-realistic output sequence.
Such a framework allows for interactive control of a portrait video.

%% file: applications/text-based-editing.tex
\subsubsection{Editing Video by Editing Text}
\label{app:text-based-editing}

\emph{Text-based Editing of Talking-head Video} \cite{Fried2019} takes as input a one-hour long video of a person speaking, and the transcript of that video. The editor changes the transcript in a text editor, and the system synthesizes a new video in which the speaker appears to be speaking the revised transcript (\Cref{fig:text-based-editing}). The system supports cut, copy and paste operations, and is also able to generate new words that were never spoken in the input video.
\input{applications/figures/text-based-editing.tex}
The first part of the pipeline is not learning-based. Given a new phrase to synthesize, the system finds snippets in the original input video that, if combined, will appear to be saying the new phrase. To combine the snippets, a parameterized head model is used, similarly to Deep Video Portraits~\cite{Kim:2018:DVP:3197517.3201283}. Each video frame is converted to a low-dimensional representation, in which expression parameters (\ie what the person is saying and how they are saying it) are decoupled from all other properties of the scene (\eg head pose and global illumination). The snippets and parameterized model are then used to synthesize a low-fidelity render of the desired output.
Neural rendering is used to convert the low-fidelity render into a photo-realistic frame. A GAN-based encoder-decoder, again similar to Deep Video Portraits, is trained to add high frequency details (e.g., skin pores) and hole-fill to produce the final result. The neural network is person specific, learning a person's appearance variation in a given environment. In contrast to  Kim~\etal~\cite{Kim:2018:DVP:3197517.3201283}, this network can deal with dynamic backgrounds.

%% file: applications/figures/text-based-editing.tex
\begin{figure}[t]
	\centering
	\includegraphics[width=\linewidth]{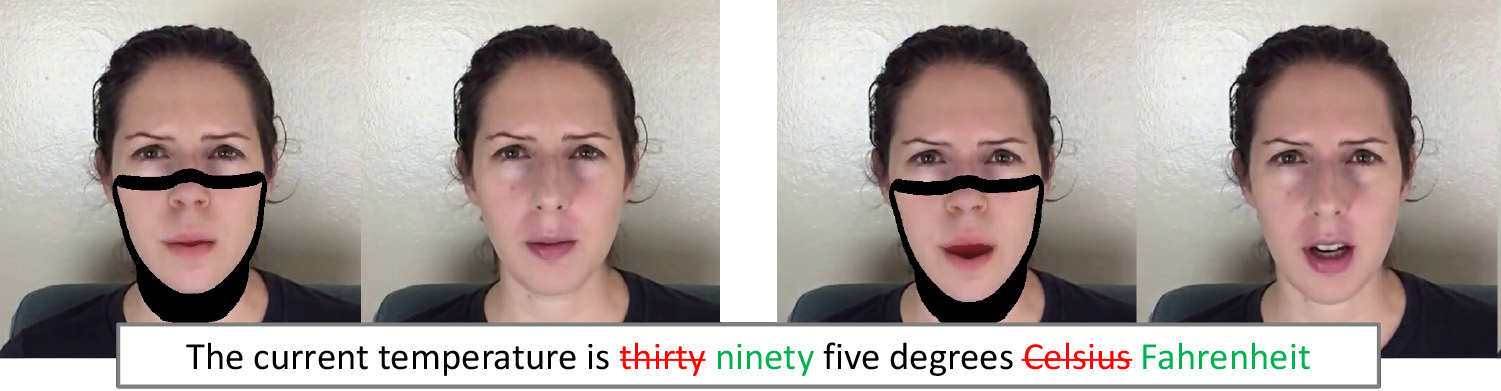}
	\vspace{-0.6cm}
	\caption
	{
	    \emph{Text-based Editing of Talking-head Video}~\cite{Fried2019}. An editor changes the text transcript to create a new video in which the subject appears to be saying the new phrase. In each pair, left: composites containing real pixels, rendered pixels and transition regions (in black); right: photo-realistic neural-rendering results.
	    \newstuff{\footnotesize{Images taken from Fried~\etal~\cite{Fried2019}}}.
	}
	\vspace{-0.5cm}
	\label{fig:text-based-editing}
\end{figure}

%% file: applications/neural-textures.tex
\subsubsection{Image Synthesis using Neural Textures}
\label{app:neuraltextures}

\textit{Deferred Neural Rendering}~\cite{Thies:2019:DNR:3306346.3323035} enables novel-view point synthesis as well as scene-editing in 3D (geometry deformation, removal, copy-move).
It is trained for a specific scene or object.
Besides ground truth color images, it requires a coarse reconstructed and tracked 3D mesh including a texture parametrization.
Instead of a classical texture as used by Kim~\etal~\cite{Kim:2018:DVP:3197517.3201283}, the approach learns a neural texture, a texture that contains neural feature descriptors per surface point.
A classical computer graphics rasterizer is used to sample from these neural textures, given the 3D geometry and view-point, resulting in a projection of the neural feature descriptors onto the image plane.
The final output image is generated from the rendered feature descriptors using a small U-Net, which is trained in conjunction with the neural texture.
The paper shows several applications based on color video inputs, including novel-view point synthesis, scene editing and animation synthesis of portrait videos. %
The learned neural feature descriptors and the decoder network compensate for the coarseness of the underlying geometry, as well as for tracking errors, while the classical rendering step ensures 3D-consistent image formation.
Similar to the usage of neural textures on meshes, Aliev et al.~\cite{Aliev2019} propose to use vertex-located feature descriptors and point based rendering to project these to the image plane.
Given the splatted features as input, a U-Net architecture is used for image generation.
They show results on objects and room scenes.

%% file: applications/talking_head_models.tex
\subsubsection{Neural Talking Head Models}
\label{app:talking_head_models}
The facial reenactment approaches we have discussed so far~\cite{Kim:2018:DVP:3197517.3201283,Fried2019,Thies:2019:DNR:3306346.3323035} are person-specific, i.e., a different network has to be trained for each identity. 
In contrast, a \textit{generalized} face reenactment approach was proposed by Zakharov~\etal~\cite{zakharov2019few}. 
The authors train a common network to control faces of any identity using sparse 2D keypoints. 
The network consists of an embedder network to extract pose-independent identity information. 
The output of this network is then fed to the generator network, which learns to transform given input keypoints into photo-realistic frames of the person. 
A large video dataset~\cite{Chung18b} consisting of talking videos of a large number of identities is used to train the network. 
At test time, few-shot learning is used to refine the network for an unseen identity\newstuff{, similar to Liu~\etal~\cite{liu2019few}.}
While the approach allows for control over unseen identities, it does not allow for explicit 3D control of scene parameters such as pose and expressions. 
It needs a reference video to extract the keypoints used as input for the network. 

%% file: applications/deep_appearance_models.tex
\subsubsection{Deep Appearance Models}
\label{app:deep_appearance_models}

\emph{Deep Appearance Models} \cite{Lombardi:2018:DAM:3197517.3201401} model facial geometry and appearance with a conditional variational autoencoder. The VAE compresses input mesh vertices and a texture map into a small latent encoding of facial expression. Importantly, this VAE is conditioned on the viewpoint of the camera used to render the face. This enables the decoder network to correct geometric tracking errors by decoding a texture map that reprojects to the correct place in the rendered image. The result is that the method produces high-quality, high-resolution viewpoint-dependent renderings of the face that runs at 90Hz in virtual reality.
The second part of this method is a system for animating the learned face model from cameras mounted on a virtual reality headset. Two cameras are placed inside the headset looking at the eyes and one is mounted at the bottom looking at the mouth. To generate correspondence between the VR headset camera images and the multi-view capture stage images, the multi-view capture stage images are re-rendered using image-based rendering from the point-of-view of the VR headset cameras, and a single conditional variational autoencoder is used to learn a common encoding of the multi-view capture images and VR headset images, which can be regressed to the latent facial code learned in the first part.
Wei~\etal~\cite{Wei:2019:VFA:3306346.3323030} improve the facial animation system by solving for the latent facial code that decodes an avatar such that, when rendered from the perspective of the headset, most resembles a set of VR headset camera images, see \Cref{fig:vrfacialanimation}. To make this possible, the method uses a ``training'' headset, that includes an additional 6 cameras looking at the eyes and mouth, to better condition the analysis-by-synthesis formulation. The domain gap between rendered avatars and headset images is closed by using unsupervised image-to-image translation techniques. This system is able to more precisely match lip shapes and better reproduce complex facial expressions.
\input{applications/figures/vrfacialanimation.tex}

%% file: applications/figures/vrfacialanimation.tex
\begin{figure}[t]
	\centering
	\includegraphics[width=\linewidth]{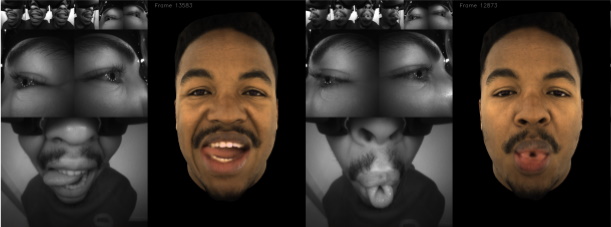}
	\caption
	{
	   Correspondence found by the system of Wei~\etal~\cite{Wei:2019:VFA:3306346.3323030}.
	   Cameras placed inside a virtual reality head-mounted display (HMD) produce a set of infrared images. Inverse rendering is used to find the latent code of a Deep Appearance Model \cite{Lombardi:2018:DAM:3197517.3201401} corresponding to the images from the HMD, enabling a full-face image to be rendered. \newstuff{\footnotesize{Images taken from Wei~\etal~\cite{Wei:2019:VFA:3306346.3323030}}}.
	}
	\label{fig:vrfacialanimation}
\end{figure}

%% file: applications/neural-bodies.tex
\subsection{Body Reenactment}
\label{app:neuralbody}
\input{applications/figures/neural-bodies.tex}
Neural pose-guided image \cite{SiaroSLS2017,ma17,Esser2018} and video  \cite{LiuBody2018,Chan2018,AbermanSLLCC19} generation
enables the control of the position, rotation, and body pose of a person in a target image/video (see \Cref{fig:neural-bodies}).
The problem of generating realistic images of the full human body is challenging, due to the large non-linear motion space of humans.
Full body performance cloning approaches \cite{LiuBody2018,Chan2018,AbermanSLLCC19} transfer the motion in a source video to a target video.
These approaches are commonly trained in a person-specific manner, i.e., they require a several-minutes-long video (often with a static background) as training data for each new target person.
In the first step, the motion of the target is reconstructed based on sparse or dense human performance capture techniques.
This is required to obtain the paired training corpus (pose and corresponding output image) for supervised training of the underlying neural rendering approach.
Current approaches cast the problem of performance cloning as learning a conditional generative mapping based on image-to-image translation networks.
The inputs are either joint heatmaps \cite{AbermanSLLCC19}, the rendered skeleton model \cite{Chan2018}, or a rendered mesh of the human \cite{LiuBody2018}.
The approach of Chan~\etal~\cite{Chan2018} predicts two consecutive frames of the output video and employs a space-time discriminator for more temporal coherence.
For better generalization, the approach of Aberman~\etal~\cite{AbermanSLLCC19} employs a network with two separate branches that is trained in a hybrid manner based on a mixed training corpus of paired and unpaired data.
The paired branch employs paired training data extracted from a reference video to directly supervise image generation based on a reconstruction loss.
The unpaired branch is trained with unpaired data based on an adversarial identity loss and a temporal coherence loss.
\emph{Textured Neural Avatars}~\cite{Shysheya2019TexturedNA} predict dense texture coordinates based on rendered skeletons to sample a learnable, but static, RGB texture.
Thus, they remove the need of an explicit geometry at training and test time by mapping multiple 2D views to a common texture map.
Effectively, this maps a 3D geometry into a global 2D space that is used to re-render an arbitrary view at test-time using a deep network.
The system is able to infer novel viewpoints by conditioning the network on the desired 3D pose of the subject.
Besides texture coordinates, the approach also predicts a foreground mask of the body.
To ensure convergence, the authors prove the need of a pre-trained DensePose model~\cite{dense_pose} to initialize the common texture map, at the same time they show how their training procedure improves the accuracy of the 2D correspondences and sharpens the texture map by recovering high frequency details.
Given new skeleton input images they can also drive the learned pipeline.
This method shows consistently improved generalization compared to standard image-to-image translation approaches.
On the other hand, the network is trained per-subject and cannot easily generalize to unseen scales.
Since the problem of human performance cloning is highly challenging, none of the existing methods obtain artifact-free results.
Remaining artifacts range from incoherently moving surface detail to partly missing limbs.

%% file: applications/figures/neural-bodies.tex
\begin{figure}[t]
	\centering
	\includegraphics[width=\linewidth]{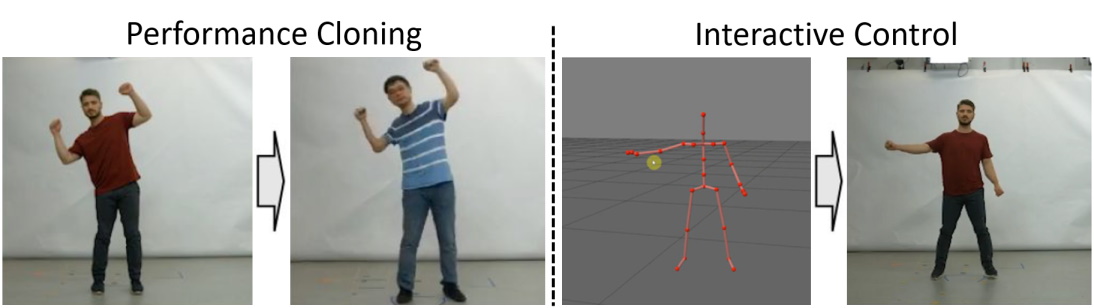}
	\caption
	{
	Neural pose-guided image and video generation enables the control of the position, rotation, and body pose of a person in a target video.
	For human performance cloning the motion information is extracted from a source video clip (left).
	Interactive user control is possible by modifying the underlying skeleton model (right).
	\newstuff{\footnotesize{Images taken from Liu~\etal~\cite{LiuBody2018}}}.
	}
	\label{fig:neural-bodies}
	\vspace{-0.5cm}
\end{figure}

%% file: 8_open_challenges.tex
\section{Open Challenges}
As this survey shows, significant progress on neural rendering has been made over the last few years and it had a high impact on a vast number of application domains.
Nevertheless, we are still just at the beginning of this novel paradigm of learned image-generation approaches, which leaves us with many open challenges, but also incredible opportunities for further advancing this field.
In the following, we describe open research problems and suggest next steps.

\noindent
\textbf{Generalization.}
Many of the first neural rendering approaches have been based on overfitting to a small set of images or a particular video depicting a single person, object, or scene.
This is the best case scenario for a learning based approach, since the variation that has to be learned is limited, but it also restricts generalization capabilities.
In a way, these approaches learn to interpolate between the training examples.
As is true for any machine learning approach, they might fail if tested on input that is outside the span of the training samples.
For example, learned reenactment approaches might fail for unseen poses \cite{Kim:2018:DVP:3197517.3201283,LiuBody2018,Chan2018,AbermanSLLCC19}.
Nevertheless, the neural rendering paradigm already empowers many applications in which the data distribution at test and training time is similar.
One solution to achieve better generalization is to explicitly add the failure cases to the training corpus, but this comes at the expense of network capacity and all failures cases might not be known a priori.
Moreover, if many of the scene parameters have to be controlled, the curse of dimensionality makes capturing all potential scenarios infeasible.
Even worse, if a solution should work for arbitrary people, we cannot realistically gather training data for all potential users, and even if we could it is unclear whether such training will be successful.
Thus, one of the grand challenges for the future is true generalization to unseen settings.
For examples, first successful steps have been taken to generalize 3D-structured neural scene representations \cite{sitzmann2019srns,hologan,nguyen2018rendernet}  across object categories.
One possibility to improve generalization is to explicitly build a physically inspired inductive bias into the network.
Such an inductive bias can for example be a differentiable camera model or an explicit 3D-structured latent space.
This analytically enforces a truth about the world in the network structure and frees up network capacity.
Together, this enables better generalization, especially if only limited training data is available.
Another interesting direction is to explore how additional information at test time can be employed to improve generalization, e.g., a set of calibration images \cite{Pandey_2019_CVPR} or a memory bank.

\noindent
\textbf{Scalability.}
So far, a lot of the effort has focused on very specific applications that are constrained in the complexity and size of the scenes they can handle.
For example, work on (re-)rendering faces has primarily focused on processing a single person in a short video clip.
Similarly, neural scene representations have been successful in representing individual objects or small environments of limited complexity.
While network generalization may be able to address a larger diversity of objects or simple scenes, scalability is additionally needed to successfully process complex, cluttered, and large scenes, for example to enable dynamic crowds, city- or global-scale scenes to be efficiently processed.
Part of such an effort is certainly software engineering and improved use of available computational resources, but one other possible direction that could allow neural rendering techniques to scale is to let the network reason about compositionality. 
A complex scene can be understood as the sum of its parts.
For a network to efficiently model this intuition, it has to be able to segment a scene into objects, understand local coordinate systems, and robustly process observations with partial occlusions or missing parts.
Yet, compositionality is just one step towards scalable neural rendering techniques and other improvements in neural network architectures and steps towards  unsupervised learning strategies have to be developed. 

\noindent
\textbf{Editability.}
Traditional computer graphics pipelines are not only optimized for modeling and rendering capabilities, but they also allow all aspects of a scene to be edited either manually or through simulation.
Neural rendering approaches today do not always offer this flexibility.
Those techniques that combine learned parameters with traditional parts of the pipeline, such as neural textures, certainly allow the traditional part (i.e., the mesh) to be edited but it is not always intuitive how to edit the learned parameters (i.e., the neural texture).
Achieving an intuitive way to edit abstract feature-based representations does not seem straightforward, but it is certainly worth considering how to set up neural rendering architectures to allow artists to edit as many parts of the pipeline as possible. 
Moreover, it is important to understand and reason about the network output as well.
Even if explicit control may not be available in some cases, it may be useful to reason about failure cases.

\noindent
\textbf{Multimodal Neural Scene Representations.}
This report primarily focuses on rendering applications and, as such, most of the applications we discuss revolve around using images and videos as inputs and outputs to a network. 
A few of these applications also incorporate sound, for example to enable lip synchronization in a video clip when the audio is edited \cite{Fried2019}.
Yet, a network that uses both visual and audio as input may learn useful ways to process the additional input modalities.
Similarly, immersive virtual and augmented reality experiences and other applications may demand multimodal output of a neural rendering algorithm that incorporates spatial audio, tactile and haptic experiences, or perhaps olfactory signals.
Extending neural rendering techniques to include other senses could be a fruitful direction of future research.

%% file: 7_social_implications.tex
\section{Social Implications}
In this paper, we present a multitude of neural rendering approaches, with various applications and target domains.
While some applications are mostly irreproachable, others, while having legitimate and extremely useful use cases, can also be used in a nefarious manner (e.g., talking-head synthesis).
Methods for image and video manipulation are as old as the media themselves, and are common, for example, in the movie industry.
However, neural rendering approaches have the potential to lower the barrier for entry, making manipulation technology accessible to non-experts with limited resources.
While we believe that all the methods discussed in this paper were developed with the best of intentions, and indeed have the potential to positively influence the world via better communication, content creation and storytelling, we must not be complacent.
It is important to proactively discuss and devise a plan to limit misuse.
We believe it is critical that synthesized images and videos clearly present themselves as synthetic.
We also believe that it is essential to obtain permission from the content owner and/or performers for any alteration before sharing a resulting video.
Also, it is important that we as a community continue to develop forensics, fingerprinting and verification techniques (digital and non-digital) to identify manipulated video (\Cref{subset:forensics}). 
Such safeguarding measures would reduce the potential for misuse while allowing creative uses of video editing technologies.
Researchers must also employ responsible disclosure when appropriate, carefully considering how and to whom a new system is released. In one recent example in the field of natural language processing \cite{radford2019language}, the authors adopted a ``staged release'' approach~\cite{solaiman2019release}, refraining from releasing the full model immediately, instead releasing increasingly more powerful versions of the implementation over a full year. The authors also partnered with security researchers and policymakers, granting early access to the full model.
Learning from this example, we believe researchers must make disclosure strategies a key part of any system with a potential for misuse, and not an afterthought.
We hope that repeated demonstrations of image and video synthesis
will teach people to think more critically about the media they consume, especially if there is no proof of origin.
We also hope that publication of the details of such systems can spread awareness and knowledge regarding their inner workings, sparking and enabling associated research into the aforementioned forgery detection, watermarking and verification systems.
Finally, we believe that a robust public conversation is necessary to
create a set of appropriate regulations and laws that would balance
the risks of misuse of these tools against the importance of creative,
consensual use cases.
For an in-depth analysis of security and safety considerations of AI systems, we refer the reader to~\cite{brundage2018malicious}.
While most measures described in this section involve law, policy and educational efforts, one measure --- media forensics --- is a \emph{technical} challenge, as we describe next.

\subsection{Forgery Detection}
\label{subset:forensics}
Integrity of digital content is of paramount importance nowadays.
\newstuff{
The verification of the integrity of an image can be done using a pro-active protection method, like digital signatures and watermarking, or a passive forensic analysis.
An interesting concept is the `Secure Digital Camera'~\cite{Blythe2004SecureDC} which not only introduces a watermark but also stores a biometric identifier of the person who took the photograph.
While watermarking for forensic applications is explored in the literature, camera manufactures have so far failed to implement such methods in camera hardware~\cite{Blythe2004SecureDC,Yan2017MultiScaleDM,Korus2015TowardsPS,Korus2018ContentAF}.
Thus, automatic passive detection of synthetic or manipulated imagery gains more and more importance.
}
There is a large corpus of digital media forensic literature which splits up in manipulation-specific and manipulation-independent methods.
\textit{Manipulation-specific detection methods} learn to detect the artifacts produced by a specific manipulation method.
FaceForensics++~\cite{roessler2019faceforensics++} offers a large-scale dataset of different image synthesis and manipulation methods, suited to train deep neural networks in a supervised fashion.
It is now the largest forensics dataset for detecting facial manipulations with over 4 million images.
In addition, they show that they can train state-of-the-art neural networks to achieve high detection rates even under different level of image compression.
Similar, Wang et al.~\cite{wang2019detecting} scripted photoshop to later detect photoshopped faces.
The disadvantage of such manipulation-specific detection methods is the need of a large-scale training corpus per manipulation method.
In ForensicTransfer~\cite{cozzolino2018forensictransfer}, the authors propose a few-shot learning approach.
Based on a few samples of a previously unseen manipulation method, high detection rates can be achieved even without a large (labeled) training corpus.
In the scenario where no knowledge or samples of a manipulation method (i.e., ``in the wild'' manipulations) are available, \textit{manipulation-independent methods} are required.
These approaches concentrate on image plausibility.
Physical and statistical information have to be consistent all over the image or video (e.g., shadows or JPEG compression).
This consistency can for example be determined in a patch-based fashion~\cite{Cozzolino2018NoiseprintAC,huh18forensics}, where one patch is compared to another part of the image.
Using such a strategy, a detector can be trained on real data only using patches of different images as negative samples.

%% file: 9_conclusions.tex
\section{Conclusion}
\label{sec:conclusions}

Neural rendering has raised a lot of interest in the past few years.
This state-of-the-art report reflects the immense increase of research in this field.
It is not bound to a specific application but spans a variety of use-cases that range from novel-view synthesis, semantic image editing, free viewpoint videos, relighting, face and body reenactment to digital avatars.
Neural rendering has already enabled applications that were previously intractable, such as rendering of digital avatars without any manual modeling.
We believe that neural rendering will have a profound impact in making complex photo and video editing tasks accessible to a much broader audience.
We hope that this survey will introduce neural rendering to a large research community, which in turn will help to develop the next generation of neural rendering and graphics applications.

%% file: 10_acks.tex
\noindent\newstuff{\footnotesize{\textbf{Acknowledgements} G.W.~was supported by an Okawa Research Grant, a Sloan Fellowship, NSF Awards IIS~1553333 and CMMI~1839974, and a PECASE by the ARL.
V.S.~was supported by a Stanford Graduate Fellowship.
C.T.~was supported by the ERC Consolidator Grant 4DRepLy (770784).
M.N.~was supported by Google, Sony, a TUM-IAS Rudolf M{\"o}{\ss}bauer Fellowship, the ERC Starting Grant Scan2CAD (804724), and a Google Faculty Award.
M.A.~and O.F.~were supported by the Brown Institute for Media Innovation.
We thank David Bau, Richard Zhang, Taesung Park, and Phillip Isola for proofreading.}}